\documentclass[twocolumn]{IEEEtran} 
\usepackage[T1]{fontenc}
\usepackage{color}
\usepackage{mathrsfs}
\usepackage{amsmath}
\usepackage{amsthm} 
\usepackage{amssymb}
\usepackage{graphicx}
\usepackage{cite}
\usepackage{makecell}
\usepackage{algorithm}
\usepackage{algorithmic}
\usepackage{fancyhdr}
\usepackage{diagbox}
\usepackage{bm}
\usepackage[unicode=true,
 bookmarks=true,bookmarksnumbered=true,bookmarksopen=true,bookmarksopenlevel=1,
 breaklinks=false,pdfborder={0 0 0},pdfborderstyle={},backref=false,colorlinks=false]
 {hyperref}
\hypersetup{
 pdftitle={Your Title},
 pdfauthor={Your Name},
 pdfpagelayout=OneColumn, pdfnewwindow=true, pdfstartview=XYZ, plainpages=false}

\usepackage{multirow} 
\usepackage{xcolor}

\allowdisplaybreaks[4]

\usepackage[caption=false,font=footnotesize]{subfig}

\IEEEoverridecommandlockouts

\usepackage{geometry}
\begin{document}
\bibliographystyle{IEEEtran}
\title{\textcolor{black}{Joint Computation Offloading and Resource Allocation for Uncertain Maritime MEC via Cooperation of UAVs and Vessels}}

\author{\IEEEauthorblockN{Jiahao You, Ziye Jia, \textit{Member, IEEE,} Chao Dong, \IEEEmembership{Senior Member,~IEEE,} Qihui Wu, \IEEEmembership{Fellow,~IEEE,} \\ and Zhu Han, \IEEEmembership{Fellow,~IEEE}}
\thanks{
  
Jiahao You, Chao Dong, and Qihui Wu are with the College of Electronic and Information Engineering, Nanjing University of Aeronautics and Astronautics, Nanjing 211106, China (e-mail: yjiahao@nuaa.edu.cn, dch@nuaa.edu.cn, wuqihui@nuaa.edu.cn). 
\textit{(Corresponding author: Chao Dong.)}

Ziye Jia is with the College of Electronic and Information Engineering, Nanjing University of Aeronautics and Astronautics, Nanjing 211106, China, and also with the National Mobile Communications Research Laboratory, Southeast University, Nanjing, Jiangsu, 211111, China (e-mail: jiaziye@nuaa.edu.cn).

Zhu Han is with the University of Houston, Houston, TX 77004 USA, and also with the Department of Computer Science and Engineering, Kyung Hee University, Seoul 446-701, South Korea (e-mail: hanzhu22@gmail.com).
}}

\maketitle
\begin{abstract}
The computation demands from the maritime Internet of Things (MIoT) increase rapidly in recent years, and the unmanned aerial vehicles (UAVs) and vessels based multi-access edge computing (MEC) can fulfill these MIoT requirements. 
However, the uncertain maritime tasks present significant challenges of inefficient computation offloading and resource allocation. 
In this paper, we focus on the maritime computation offloading and resource allocation through the cooperation of UAVs and vessels, with consideration of uncertain tasks.
Specifically, we propose a cooperative MEC framework for computation offloading and resource allocation, including MIoT devices, UAVs and vessels. 
Then, we formulate the optimization problem to minimize the total execution time. 
As for the uncertain MIoT tasks, we leverage Lyapunov optimization to tackle the unpredictable task arrivals and varying computational resource availability.	
By converting the long-term constraints into short-term constraints, we obtain a set of small-scale optimization problems.
Further, considering the heterogeneity of actions and resources of UAVs and vessels, we reformulate the small-scale optimization problem into a Markov game (MG).
Moreover, a heterogeneous-agent soft actor-critic is proposed to sequentially update various neural networks and effectively solve the MG problem.	
Finally, simulations are conducted to verify the effectiveness in addressing computational offloading and resource allocation.
\end{abstract}


\section{Introduction}\label{s1}
\IEEEPARstart{I}{n} recent years, the increasing demands for global trade, resource exploration, fisheries, and tourism drive the expansion of maritime activities \cite{Maritime_2020}.	
Then, the maritime Internet of Things (MIoT) employs sensors and wireless networks to collect, transmit, analyze data, and enhance the intelligence of maritime management.
However, most MIoT devices remain unexplored, due to complex and uncertain marine environment \cite{Zhu_Software_2022, Internet_2021, Joint_2024}.		
The MIoT devices face challenges such as varied communication environments, computational constraints, and limited energy resources.	
Some applications necessitate high data rates and low delay, particularly for sea lane monitoring and navigation assistance, which require real-time data processing and analysis abilities.	
Nevertheless, the limited communication and computing resources for maritime services present significant challenges. 
Fortunately, the multi-access edge computing (MEC) brings computing capabilities of UAVs and vessels to the network edge, and allows for faster data processing and reduced delay by bringing computation and storage closer to the data sources.
Therefore, it is essential to integrate MEC technologies, leveraging the capabilities of UAVs and vessels \cite{dong_uavs_2021,Cooperative_2024}. 
Such integration can optimize computation offloading and resource allocation, and enhance the efficiency and reliability of maritime services.

The UAV-assisted MEC provides an effective solution for MIoT devices to manage compute-intensive tasks by deploying computing resources at the network edge \cite{ yang_multi-armed_2022, zhao_multi-agent_2022}.	
However, the UAV based MEC faces restricted computational capacity \cite{Energy_2024}, energy \cite{PDDQNLP_2023}, and transmission power \cite{Computation_2024, CIntelligent_2024}. 
Hence, it is challenging for UAVs to handle large MIoT data independently, and some tasks may encounter unacceptable delay \cite{Latency_2024}.	
Therefore, the cooperation between vessels and UAVs is an effective mechanism, in which vessels can provide more substantial computational resources and higher energy capacity.
In detail, UAVs can handle a portion of MIoT tasks, typically involving data collection and preprocessing, and relay the computation-intensive tasks to vessels for powerful MEC services \cite{Task_2024_Dai, wang_hybrid_2021,Jia_LEO_2021}.	
However, the integration of heterogeneous resource devices is challenging due to varied capabilities and operational environments.	
Additionally, the uncertainty in maritime tasks complicates the resource management. Hence, the computation offloading and resource allocation in maritime environments face the following challenges.	
\begin{itemize}
\item In maritime scenarios, computational tasks are uncertain due to unpredictable task arrivals and dynamic resource availability, which increase the complexity of computation offloading and resource allocation.
\item The cooperation between UAVs and vessels offers potential for enhanced MEC, but the heterogeneous devices bring challenges for cooperated resource allocation.
\end{itemize}

To overcome above challenges, we investigate the maritime MEC cooperated by UAVs and vessels. 
In detail, we propose a cooperative MEC framework for maritime network. Then, we formulate a resource allocation and computation offloading problem to minimize the total execution time, which is in the form of mixed-integer program (MIP) and NP-hard to solve, especially in large scale \cite{Shuai_Transfer_2022}.
Considering the uncertain maritime tasks, we utilize the Lyapunov optimization to transform the original problem into a set of per-time-slot small-scale optimization problems.
Subsequently, considering the heterogeneity of actions and resources of UAVs and vessels in the MEC environment, we reformulate the small-scale optimization problem as a Markov game (MG).
Then, we design a heterogeneous-agent soft actor-critic (HASAC) algorithm to handle the MG problem. The main contributions of this paper are summarized as follows.
\begin{itemize}
  \item We propose a cooperative MEC framework for maritime MEC, in which MIoTs utilize the computational resources of UAVs and vessels. Besides, The framework considers task arrival times and the dynamic availability of computational resources under queue stability constraints.
  \item To address the uncertainty, we employ Lyapunov optimization to transform the long-term constraints on the execution time into per-time-slot small-scale problems. To handle the significant challenge of integrating heterogeneous resource devices, we reformulate the small-scale problem as a MG.
  \item We design the HASAC algorithm to solve the problem of uncertainty and heterogeneity in maritime environment by combining soft policy iteration and reinforcement learning, to tackle the MG problem arising from the varied task demands and resource capabilities of UAVs and vessels.
\end{itemize}

The rest of the paper is organized as follows. 
Related works are discussed in Section \ref{s2}.	
The system model is detailed in Section \ref{s3}, followed by the problem formulation in Section \ref{s4}.	
Subsequently, Section \ref{s5} introduces the HASAC algorithm.	
Numerical results are presented in Section \ref{s6}, with conclusions drawn in Section \ref{s7}.	

\section{Related Work}\label{s2}
\subsection{UAV-assisted Maritime MEC}
In the maritime network, the integration of MEC is essential to enhance the system performance by providing low-delay, high-bandwidth services, and efficient resource management. 
For example, Liu \textit{et al.} \cite{Deep_Maritime_2022} design a two-layer UAV-enabled MEC network to optimize wireless network performance and minimize delay in maritime environments. They utilize reinforcement learning to optimize UAV trajectory and virtual machine configuration, to tackle the challenges of resource scarcity and delay sensitivity in maritime MEC.
Due to the vulnerability of the line-of-sight communication channels in maritime networks, Lu \textit{et al.} \cite{Resource_Maritime_2024} introduce a secure communication scheme for UAV-relay-assisted maritime MEC networks.
To address the issue of energy efficiency in UAV-assisted maritime MEC networks, \cite{Joint_Maritime_2023} proposes a non-orthogonal multiple access-based MEC model and a two-layered algorithm to jointly optimize UAVs resource allocation and UAV trajectory.	
Dai \textit{et al.} \cite{Latency_2023} propose a hybrid offshore and aerial-based MEC scheme, to handle the real-time data processing challenges in marine communication networks. 
Xu \textit{et al.} \cite{Latency_2024_ear} design an optimal offloading scheme using UAV-assisted maritime MEC, dynamically optimizing UAV trajectory, user scheduling, and resource allocation to minimize task execution time and improve the quality of service.	
However, these works do not focus on maximizing long-term network performance and uncertainty. 
In maritime MEC, the design of strategies should consider real-time uncertainties of environment, such as unpredictable task arrivals and dynamic availability of computational resources.	
Additionally, due to geographical constraints, the maritime communication and computing resources are limited, presenting challenges for resource allocation and computation offloading in maritime MEC.	

\subsection{Uncertainty in MEC}
The uncertainty reflects real-world unpredictability across a range of factors such as task arrival unpredictability, network condition instability, and task processing uncertainty.
To tackle the task arrival unpredictability, Fan \textit{et al.} \cite{Robust_Fan_2023} develop a novel resource allocation optimization method in MEC for uncertain computation loads, ensuring robust and efficient solutions with low-complexity algorithms. 
Mao \textit{et al.} \cite{Energy_2023_Mao} propose the development of a successive convex approximation-based algorithm, to optimize various network parameters in the presence of channel uncertainties in MEC. 
In addition, Xia \textit{et al.} \cite{Distributed_2023_Xia} investigate the coordination between MEC and ultra-dense networks for Internet of Things, and propose algorithms based on game theory and stochastic programming to manage computation offloading and power distribution uncertainties. 
To address the uncertainty associated with task processing, Li \textit{et al.} \cite{Enhancing_2023_Li} present processing uncertainties in MEC computation offloading by formulating a chance-constrained program.
Liu \textit{et al.} \cite{Distributed_2024} introduce game-theoretic and stochastic programming algorithms to achieve adaptive task offloading and power management, effectively handling user mobility and resource uncertainty.
Ma et al. \cite{Video__2024} tackle the challenges of video offloading in MEC under uncertainty due to dynamic device mobility, by proposing two uncertainty-aware approaches to enhance video quality and reduce service migration costs.	
However, these works do not study the uncertainties in the maritime environment.	
Hence, the factors such as the unpredictability of task arrivals and the dynamic nature of computational resources make the impact of uncertainty even more complex in the maritime environment.	

\subsection{Multi-agent Reinforcement Learning (MARL) in MEC}
The complexity and dynamic nature of space-air-ground integrated network, characterized by evolving network conditions, varied user demands, and multiple resource limitations, necessitate innovative approaches.	
MARL leverages network entities including satellites, UAVs, and ground stations as agents, enhancing learning and decision-making \cite{Single_2021_Feriani, Applications_2022_Li}.	
Li \textit{et al.} \cite{Applications_2022_Li} explore MARL applications in emerging networks such as MEC, discussing the challenges, methodologies, and potentials to resolve complex issues.	
In the context of collaborative computation offloading and service caching in multi-cell MEC networks, Yao \textit{et al.} \cite{Cooperative_2023_Yao} present a graph attention-based MARL algorithm, leveraging digital twin technology to boost simulations and analyses.	
Peng \textit{et al.} \cite{Multi_2021_Peng} propose a multi-agent deep deterministic policy gradient method for resource management in UAV-assisted vehicular networks, aiming for efficient allocation and offloading.	
Gao \textit{et al.} \cite{Large_2023_Gao} offer a decentralized attention-weighted recurrent multi-agent actor-critic solution for computation offloading in large-scale heterogeneous MEC, targeting enhanced task completion and reduced costs.
These algorithms tackle various aspects of resource allocation and computation offloading in complex network environments. 
However, it lacks consideration for the heterogeneity of agents in maritime environments, which leads to suboptimal performance in real-world applications.

\section{System Model}\label{s3}
\begin{figure}[!t]
  \centering
  \includegraphics[width=8cm]{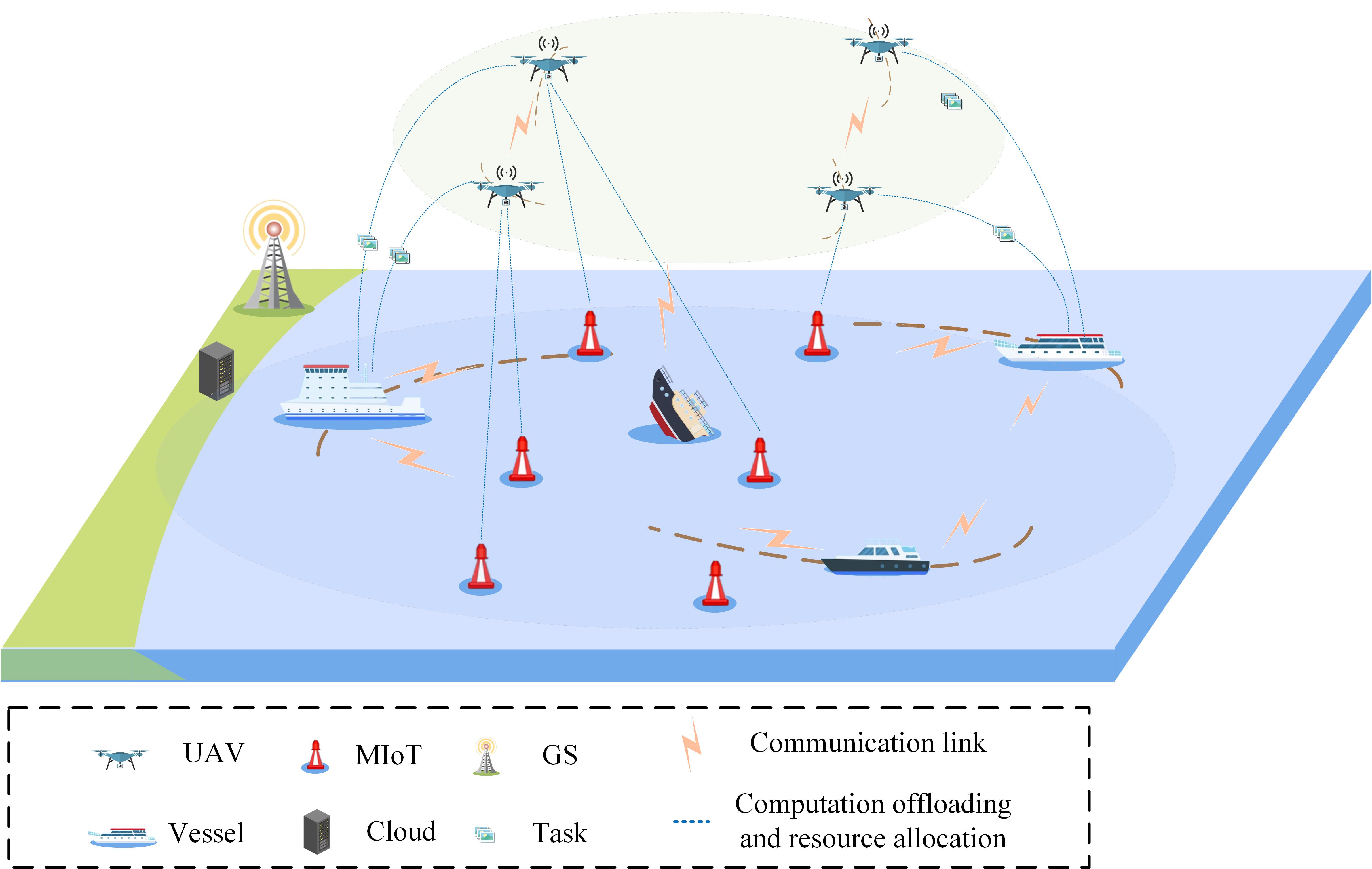}
  \caption{System overview.}
  \label{f1}
\end{figure}

As shown in Fig. \ref{f1}, the cooperative MEC framework consisting of UAVs, vessels, MIoT devices, and ground stations (GSs) is provided. 
In detail, GSs primarily transmit task commands to vessels and UAVs, and vessels are equipped with multiple receiving antennas and have powerful computing capabilities.	
Consequently, vessels can serve as computing devices for computation offloading and resource allocation. 
MIoT devices generate multiple tasks requiring computation.	
However, the limited computing and energy ability of MIoT devices may not complete the local computation.
Therefore, UAVs collect data from MIoT devices for lightweight computing tasks, and can relay compute-intensive tasks to vessels. 
This hierarchical structure efficiently manage the distribution of computational demands between UAVs and vessels.

In detail, the system components comprise MIoTs, UAVs, and vessels.	
MIoTs, UAVs, and vessels are denoted by $\mathcal{M}=\{1, 2, 3, ..., I\}$, $\mathcal{U}=\{1, 2, 3,..., J\}$, and $\mathcal{V}=\{1, 2, 3,..., K\}$, respectively.	
Vessels are equipped with $N_k^{max}$ antennas to facilitate simultaneous communication with $N_k^{max}$ UAVs.	
Then, the total time $\mathcal{T}$ is segmented into $T$ periods, each spanning a duration of $\tau$.	
Additionally, the positions of MIoTs, UAVs, and vessels are represented in a three dimensional cartesian coordinate system, i.e., $l_i^m(t) = (x_i^{m}(t), y_i^{m}(t), h_i^m(t))$, $l^u_j(t) = (x_j^{u}(t), y_j^{u}(t),h_j^u(t))$, and $l^v_k(t) = (x_k^{v}(t), y_k^{v}(t), h_k^v(t))$, respectively.
At time slot $t$, MIoT $i$ generates tasks $A_i(t) = \{d_i(t), c_i(t)\}$, where $d_i(t)$ represents the data size in bits and $c_i(t)$ indicates the required computation resources in CPU cycles per bit. To more accurately model the arrival of random events over time in real-world scenarios, we utilize a Poisson distribution for generating MIoT workloads, with a mean value of $\mathbb{E}(A_i(t)) = \lambda_i$. It is assumed that at each time slot $t$, task $A_i(t)$ arrives randomly at MIoT $i$ for offloading.

\begin{figure*}[htbp]
  \centering
  \captionsetup[subfloat]{justification=raggedright, singlelinecheck=false}
  \subfloat[Tasks are generated from MIoTs.]{
  \includegraphics[width=8cm]{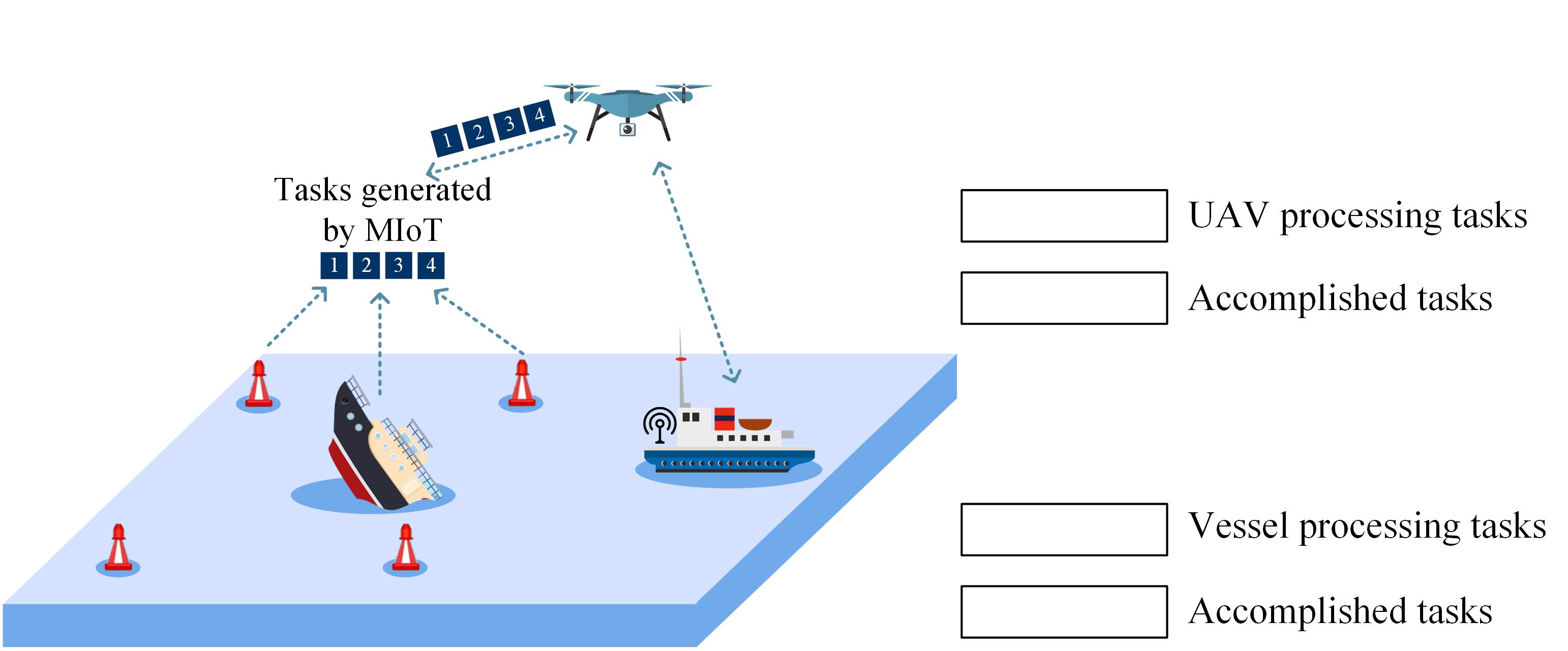}
  }
  \quad
  \subfloat[Task 1 is being processed on the UAV. Task 2 is offloaded to the vessel for processing.]{
  \includegraphics[width=8cm]{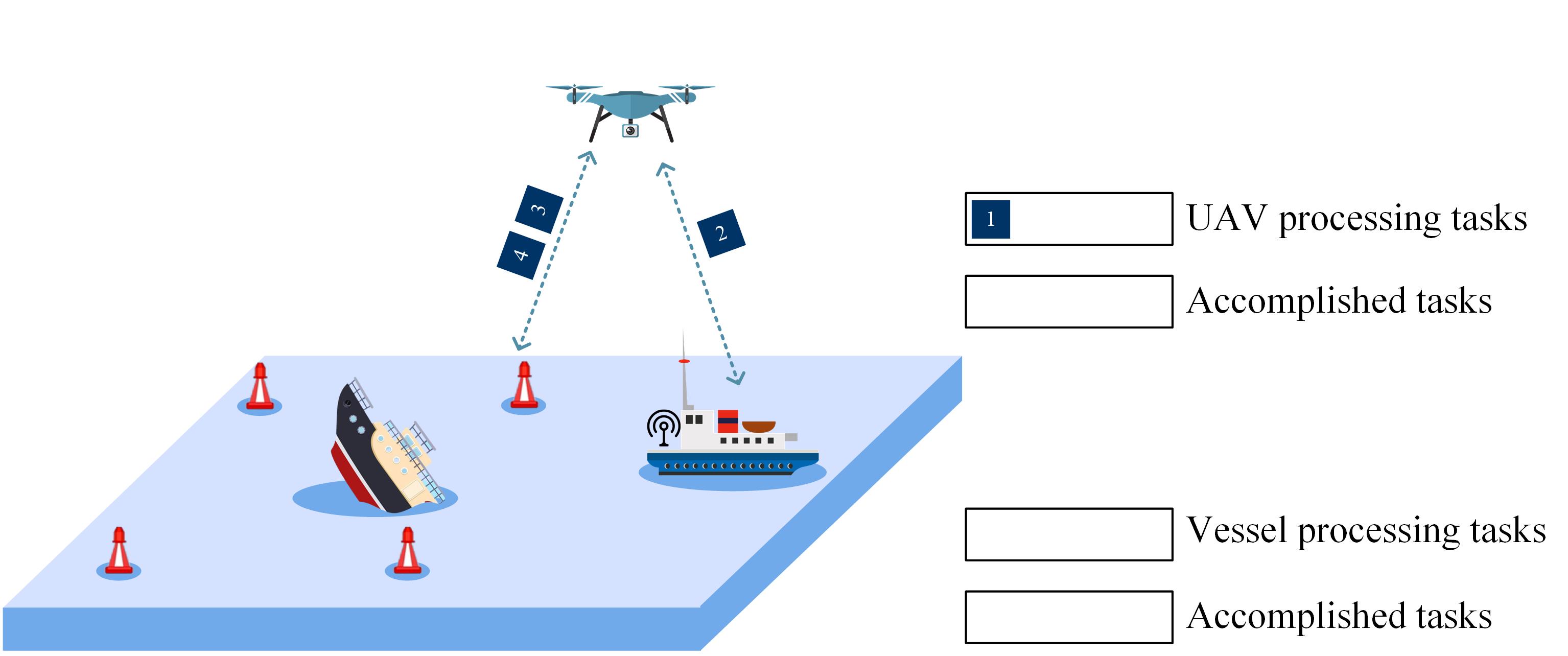}
  }
  \quad
  \subfloat[Task 1 is completed and transmitted back to MIoT. Task 2 is being processed on the UAV. Task 3 is offloaded to the vessel for processing. Task 4 is being processed on the UAV.]{
  \includegraphics[width=8cm]{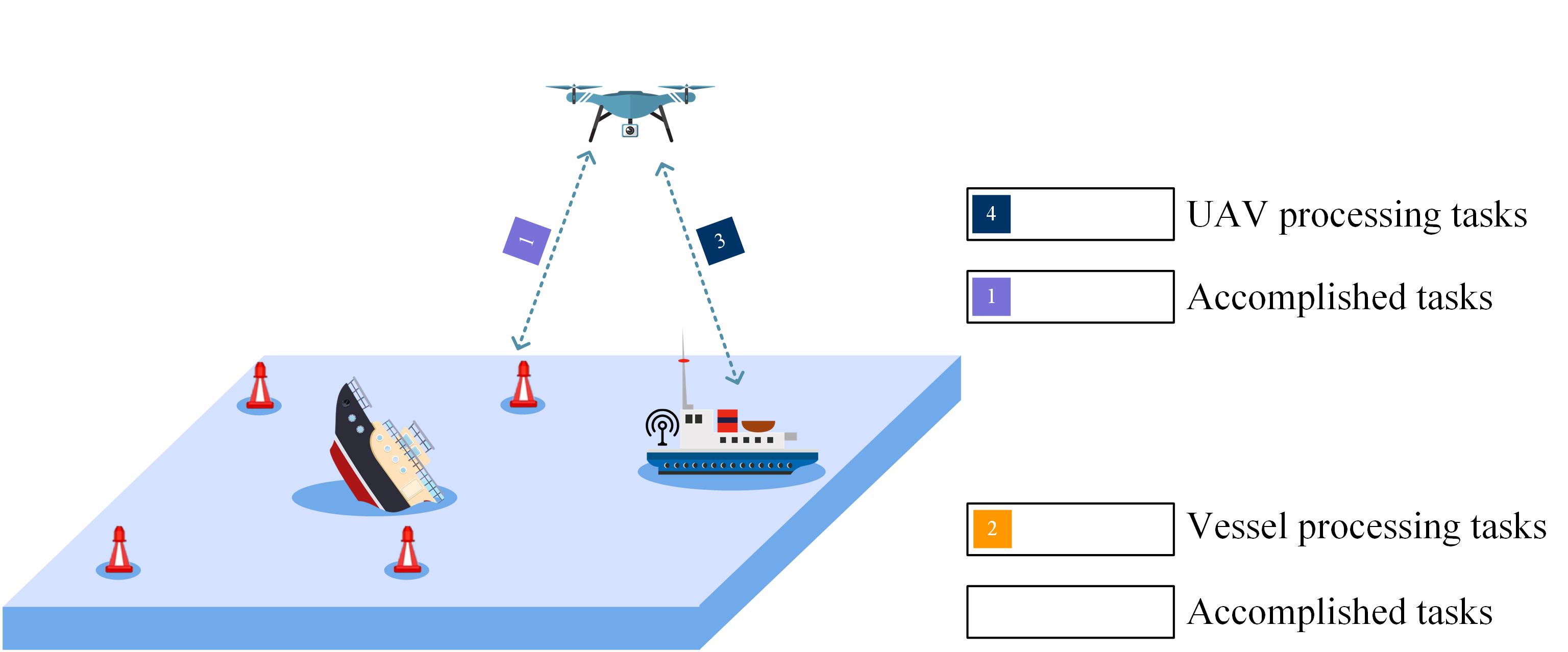}
  }
  \quad
  \subfloat[Task 2 is completed and transmitted back to UAV. Task 3 is being processed on the vessel. Task 4 completion is returned to the MIoT.]{
  \includegraphics[width=8cm]{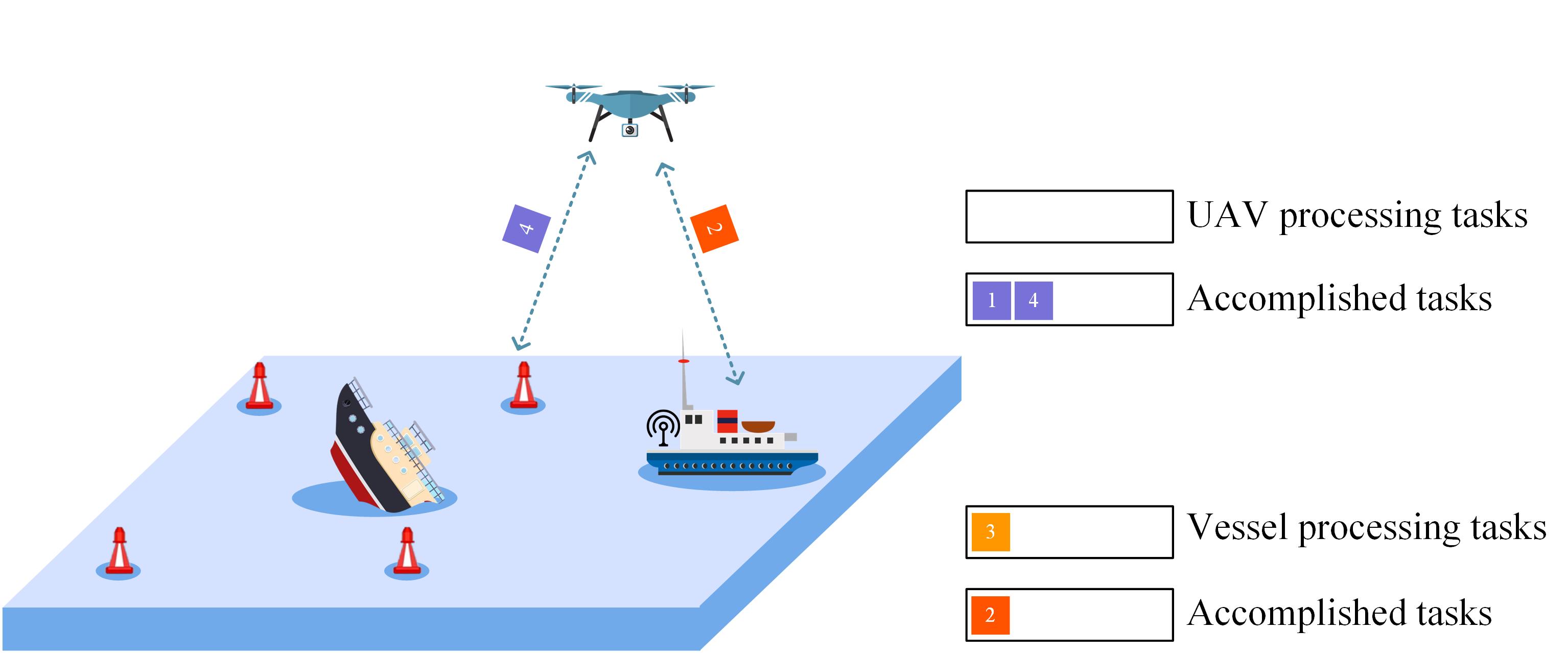}
  }
  \caption{An example of computation offloading and resource allocation.}\label{f2}
\end{figure*}

The system operates within discrete time slots, with tasks generated by MIoT devices designated for offloading to either UAV $j$ or vessel $k$.	
In addition, UAV $j$ evaluates its operational capacity, current load, resource availability, and connectivity quality with MIoTs and vessels before offloading.	
It is worth noting that the availability of computational resources is dynamic, driven by the random arrival of tasks and the mobility of UAVs. Task arrivals lead to fluctuating computational demands, while the UAV movement affects their location, computational capacity, and connectivity with MIoTs.
Based on this evaluation, UAV $j$ proactively issues an offloading request to MIoT $i$.	
When receiving a task, UAV $j$ determines whether it has sufficient computation resources for processing.	
If the computation resources are sufficient, UAV $j$ processes the task, and otherwise it offloads the task to vessel $k$. 
In the system, UAVs and vessels communicate to exchange crucial information regarding task offloading, resource availability, and status updates. UAVs and vessels share their location data, computational capacity, and current task information to ensure optimal decision-making. UAVs transmit their available resources and offloading decisions to vessels, while vessels provide feedback on resource allocation and computational capacity.
The example shows in Fig. \ref{f2}.

It is noted that UAVs operate in a full-duplex mode, allowing for simultaneous task uploading and downloading, with the uploading time being significantly greater than the downloading time \cite{Integrating_2021, Wireless_2022}. As a result, the task backhaul time, which is primarily associated with the downloading process, becomes negligible in comparison to the uploading time and does not significantly impact the overall system performance.

\begin{table*}[h]
\caption{Notations Used Throughout The Paper}
\begin{center}
\begin{tabular}{|m{3.8cm}<{\centering}|m{11cm}|}
\hline
\textbf{Symbol} &\textbf{Description}\\
\hline
$\mathcal{M}, \mathcal{U}, \mathcal{V}$&Set of MIoTs, UAVs and vessels.\\
\hline
$i,j,k$&Index of MIoTs $\mathcal{M}$, UAVs $\mathcal{U}$ and vessels $\mathcal{V}$.\\
\hline
$\mathcal{T}, T, t, \tau$&The total time, total number of time slot, index of time slot and a single time slot length.\\
\hline
$l_i^m(t), l^u_j(t), l^v_k(t)$ &Location of MIoTs, UAVs and vessels.\\
\hline
$d_i(t), c_i(t)$ & Data size and the amount of computation of the task $A_i(t)$.\\
\hline
$\xi_{i,j}(t)$&Path loss from MIoT $i$ to UAV $j$.\\
\hline
$\zeta_{L}, \zeta_{NL}, \alpha, \beta$&Environmental parameters of MIoT $i$ to UAV $j$.\\
\hline
$R_{i,j}^{m2u}(t), R_{j,k}^{u2v}(t)$&The transmission rate between MIoT $i$ and UAV $j$, and the transmission rate from UAV $j$ to vessel $k$.\\
\hline
$G_{j,k}(t)$&The channel power gain between UAV $j$ to vessel $k$ at time slot $t$.\\
\hline
$A_i(t)$& The task arrived by MIoT $i$ in time slot $t$.\\
\hline
$Q_i^{m}(t), Q_j^{u}(t), Q_k^{v}(t)$ &The backlog of MIoT $i$, UAV $j$ and vessel $k$ in time slot $t$.\\
\hline
$T^u(t),T^v(t)$ &The delay of tasks of UAV $j$ in time slot $t$ and the delay of tasks of vessel $k$ in time slot $t$.\\
\hline
$\Phi (t)$&The time total cost of all tasks generated by MIoTs in time slot $t$.\\
\hline
$o_{i,j}(t), s_{j,k}(t)$&Offloading decision from MIoT $i$ to UAV $j$ and offloading decision from UAV $j$ to vessel $k$.\\
\hline
$f_{i,j}^u(t), f_{j,k}^v(t)$ & The computing resource allocated by UAV $j$ to MIoT $i$ in time slot $t$ and vessel $k$ to UAV $j$ in time slot $t$.\\
\hline
$\mathcal{O},\mathcal{S}$ & Set of offloading decision $o_{i,j}(t)$ and $s_{j,k}$.\\
\hline
$\mathcal{F}^u,\mathcal{F}^v$ &Set of computational resource  allocation $f_{i,j}^u(t)$ and $f_{j,k}^v(t)$.\\
\hline
\end{tabular}\label{T1}
\end{center}
\end{table*}

\subsection{Communication Model}
In this study, the considered MIoT devices are assumed to be deployed on or above the sea surface, such as buoys or surface sensors.

\subsubsection{MIoT to UAV (M2U)}
The path loss from MIoT $i$ to UAV $j$ is modeled as
\begin{equation}
\begin{split}
\xi_{i,j}(t) &= \frac{\zeta_{L}-\zeta_{NL}}{1+\alpha \exp \{-\beta [\gamma_{i,j}(t) -\alpha \}} \\
& + 20 \lg\left(\frac{4\pi\|l_i^m(t)-l_j^u(t)\| \varphi_c}{C_0}\right)+\zeta_{NL},
\end{split}
\end{equation}

where
\begin{equation}
\begin{aligned}
\gamma_{i,j}(t) = \arctan \left(\frac{\vert h_i^m-h_j^u \vert }{\|l_i^m(t)-l_j^u(t)\|_2 }\right).
\end{aligned}
\end{equation}
$\varphi_c$ denotes the carrier frequency, and $C_0$ represents the speed of light.
$\zeta_{L}$, $\zeta_{NL}$, $\alpha$, and $\beta$ are parameters characterizing the environment \cite{wang_hybrid_2021, Jia_Joint_HAP_2021}.

Utilizing the Shannon formula, the average rate between MIoT $i$ and UAV $j$ is computed as	
\begin{equation}
R_{i,j}^{m2u}(t)=o_{i,j}(t)B_0 \log_2  \left(1+\frac{P_{i}^m\xi_{i,j}(t)}{N_G}\right),
\end{equation}
where $o_{i,j}(t) = 1$ denotes that the task of MIoT $i$ is offloaded to UAV $j$ at time slot $t$, and 0 otherwise. $B_0$ is the bandwidth of the channel, $P_{i}^m$ represents the transmitting power of MIoT $i$, and $N_G$ indicates the power of the additive white Gaussian noise.

\subsubsection{UAV to Vessel (U2V)}
Let $L$ denote the number of orthogonal licensed channels provided by UAVs, each with bandwidth $B$.	
Then, the time-varying channel power gain from UAV $j$ to vessel $k$ for time slot $t$ is expressed as
\begin{equation}
G_{j,k}(t) = G_0(\|l_j^u(t)-l_k^v(t)\|_2)^{-2},
\end{equation}
where $G_0$ is the channel power gain when $|l_i^m(t)-l_j^u(t)|_2$ equals 1-m \cite{Zhou_CRM_2018,Luo_DRL_TP_2023, Jia-Hierarchical_2022}. 
Consequently, the transmission rate from UAV $j$ to vessel $k$ is 
\begin{equation}
R_{j,k}^{u2v}(t) = \frac{s_{j,k}(t)LB}{\sum_{j \in J}s_{j,k}(t)}log_2\left(1+\frac{P_{j,k}^{u}G_{j,k}(t)}{N_G}\right),
\end{equation}
where $P_{j,k}^{u}$ denotes the transmission power of UAV $j$. 
Variable $s_{j,k} = 1$ indicates that UAV $j$ offloads a task to vessel $k$ at time slot $t$, and 0 otherwise.

\subsection{Computation Model}
The computation offloading decision for MIoT $i$ at time slot $t$ is defined as
\begin{equation}
o_{i,j}(t)= \begin{cases}1, & \text {if MIoT } i \text { offloaded to UAV } j, \\ 0, & \text {otherwise. }\end{cases}\label{c2}
\end{equation}

Each MIoT is limited to offloading its tasks to a single UAV, i.e.,
\begin{equation}
\sum_{j = 1}^J o_{i,j}(t) \leq  1,\forall i \in I, t \in T.
\end{equation}

For UAV $j$ at time slot $t$, the computation offloading decision is defined as
\begin{equation}
s_{j,k}(t)= \begin{cases}1, & \text {if UAV } j \text { offloaded to vessel } k,\\ 0, & \text {otherwise. }\end{cases}
\end{equation}

It is crucial that each UAV is permitted to offload tasks to only one vessel per time slot, i.e.,
\begin{equation}
\sum_{k = 1}^K s_{j,k}(t) \leq 1, \forall j \in J, t \in T. \label{c5}
\end{equation}

Hence, the backlog of MIoT $i$ can be represented as
\begin{equation}
Q_i^{m}(t+1) = \text{max}\bigg\{Q_i^{m}(t)-\sum_{j = 1}^J \tau R_{i,j}^{m2u}(t)+ A_i(t), 0\bigg\},
\end{equation}
where $\sum_{j = 1}^J \tau R_{i,j}^{m2u}(t)$ indicates the amount of tasks leaving MIoT $i$ in time slot $t$. At time slot $0$, let $Q_i^{m}(0) = 0$. 

\subsubsection{UAVs based Computing}
$f_j^{umax}$ is defined as UAV $j$ total computing resource, with $f_{i,j}^u(t)$ denoting the resource allocated to MIoT $i$ during time slot $t$.
Thus, task delay in time slot $t$ includes both transmission and computation latencies, calculated as
\begin{equation}
T^u(t) = \frac{d_i(t)}{R_{i,j}^{m2u}(t)} +\frac{d_i(t) c_i(t)}{f_{i,j}^u(t)}.
\end{equation}

Furthermore, the computing resources allocated by UAV $j$ to MIoT devices cannot exceed its total resources, i.e.,
\begin{equation}
\sum_{i = 1}^I f_{i,j}^u(t) \leq  f_j^{umax}, \forall j \in J, t \in T. \label{c7}
\end{equation}

Besides, UAV $j$ maintains a queue for tasks offloaded by MIoT $i$, with the backlog modeled as $Q_i^{u}(t)$, i.e.,
\begin{equation}
\begin{split}
Q_j^{u}(t+1) = \text{max}\bigg\{&Q_j^{u}(t)- \sum_{k = 1}^K \tau R_{j,k}^{u2v}(t)\\
&-\tau f_{i,j}^u(t)+  \sum_{i = 1}^I \tau R_{i,j}^{m2u}(t), 0 \bigg\},
\end{split}
\end{equation}
where $\sum_{k = 1}^K \tau R_{j,k}^{u2v}(t)$ means UAV $j$ select vessel $k$ for offloading in time slot $t$, $\tau f_{i,j}^u(t)$ indicates the amount of data processed of MIoT $i$ by UAV $j$ in time slot $t$.

\subsubsection{Vessels based Computing}
$f_k^{vmax}$ is defined as the computing resource of vessel $k$, and $f_{j,k}^v(t)$ represents the resource allocated to UAV $j$ task during time slot $t$. 
Besides, computation offloading to vessels includes three types of delay: transmission from MIoTs to UAVs, propagation from UAVs to vessels, and vessel computation.	
Hence, the total task delay is represented as
\begin{equation}
  T^v(t) = \frac{d_i(t)}{R_{i,j}^{m2u}(t)} + \frac{d_i(t)}{R_{j,k}^{u2v}(t)} + \frac{d_i(t)c_i(t)}{f_{j,k}^v(t)}.
\end{equation}

The vessel total allocated computing resources for each UAV must not exceed its capacity, i.e., 
\begin{equation}
\sum_{j = 1}^J f_{j,k}^v(t) \leq  f_k^{vmax}, \forall k \in K, t \in T. \label{c9}
\end{equation}

Therefore, the tasks offloaded from UAVs are queued in the vessel server for computation.	
Then, we define the length of the task queue from UAVs offloading to vessel $k$ as
\begin{equation}
\begin{split}
Q_k^{v}(t+1) = \text{max}\bigg\{Q_k^{v}(t)-\tau f_{j,k}^v(t)+\sum_{j = 1}^J \tau R_{j,k}^{u2v}(t), 0 \bigg\},
\end{split}
\end{equation}
where $\tau f_{j,k}^v(t)$ denotes the size of data computed by the vessel servers at time slot $t$, and $\tau R_{j,k}^{u2v}(t)$ represents the amount of task data offloaded from UAV $j$ to vessel $k$ at time slot $t$.
In addition, tasks offloaded to UAVs or vessels are processed following a first-come, first-served basis.

Derived from the preceding model, tasks can be processed on UAVs or vessels.	
Hence, the delay for MIoT $i$ tasks in time slot $t$ is represented as
\begin{equation}
T_i^a(t) = T^u(t) + T^v(t).
\end{equation}

Consequently, the cumulative time cost for all MIoT-generated tasks in time slot $t$ is 
\begin{equation}
\Phi (t) = \sum_{i = 1}^{I} T_i^a(t).
\end{equation}

Then, the average time cost for all tasks is defined as 
\begin{equation}
\bar{\Phi} = \lim_{T\rightarrow\infty} \frac{1}{T}\sum_{t=1}^{T}\mathbb{E} [\Phi (t)]. 
\end{equation}

\section{Problem Formulation}\label{s4}
The objective is to minimize the total execution time during computation offloading and resource allocation.	
As previously discussed, the computation offloading and resource allocation are organized into four distinct sets.
\begin{itemize}
\item $\mathcal{O}$: Computation offloading decisions from MIoTs to UAVs;
\item $\mathcal{S}$: Computation offloading decisions from UAVs to vessels;
\item $\mathcal{F}^u$: Computing resources allocated by UAVs to MIoTs;
\item $\mathcal{F}^v$: Computing resources allocated by vessels to UAVs.
\end{itemize}
Consequently, the optimization problem is formulated as 
\begin{subequations}
\begin{align}
\mathscr{P}0: &\min_{\mathcal{O},\mathcal{S},\mathcal{F}^u,\mathcal{F}^v}\lim_{T\rightarrow\infty} \frac{1}{T}\sum_{t=1}^{T}\mathbb{E} [\Phi (t)], \notag\\
&\text{s.t. } (\ref{c2})-(\ref{c5}), (\ref{c7}), (\ref{c9}) \notag\\
& \lim_{T \rightarrow \infty} \frac{1}{T}  \sum_{t=1}^T \frac{Q_i^m(t)}{t} = 0, \forall i \in \mathcal{M},  \label{Za} \\
& \lim_{T \rightarrow \infty} \frac{1}{T}  \sum_{t=1}^T \frac{Q_j^u(t)}{t} = 0, \forall j \in \mathcal{U},\label{Zb} \\
&\lim_{T \rightarrow \infty} \frac{1}{T} \sum_{t=1}^T \frac{Q_k^v(t)}{t} = 0, \forall k \in \mathcal{V}. \label{Zc} 
\end{align}
\end{subequations}
Wherein, constraints (\ref{Za})-(\ref{Zc}) aim to ensure queue stability.	
Then, a strongly stable data queue is desirable since it ensures finite processing delay of each task \cite{book_Stochastic_Network_Optimization_2010,Zhou_Lyapunov-Guided_2021}. 
The solution of problem $\mathscr{P}0$ is a significant challenge, since the  short-term decisions impact long-term queuing delay performance, necessitating decisions making without future information.	

\subsection{Problem Transformation}
The Lyapunov optimization is employed to address the long-term stochastic problem.	
Then, we convert the long-term queuing delay constraints into queue stability constraints by introducing virtual queues \cite{Lyapunov_Wu_2023}. 
Let $Q(t) = \{Q_i^{m}(t), Q_j^{u}(t), Q_k^{v}(t)\}$ represent the set of all queues.	
Hence, we define the quadratic Lyapunov function $L(Q(t))$ as
\begin{equation}
L(Q(t)) = \frac{1}{2}\sum_{i=1}^{I}[Q_i^{m}(t)]^2+\frac{1}{2}\sum_{j=1}^{J}[Q_j^{u}(t)]^2+ \frac{1}{2}\sum_{j=1}^{J}[Q_k^{v}(t)]^2,
\end{equation}
where $L(Q(t))$ represents the system total queue congestion.	

A large backlog in the queue results in a significantly high $L(Q(t))$ value.	
Subsequently, the one-step conditional Lyapunov drift  $\Delta(Q(t))$ is defined as
\begin{equation}
\Delta(Q(t)) = \mathbb{E} [L(Q(t+1)) - L(Q(t))|Q(t)].
\end{equation}

Ensuring system queue stability necessitates a low Lyapunov function value.	
To minimize the total execution time  while maintaining queue stability, the drift-plus-penalty for each time slot is $\Delta(Q(t)) + V\mathbb{E}[\Phi (t)|Q(t)]$,
where $V$ denotes the trade-off between the total execution time and Lyapunov drift. 
In detail, a higher $V$ value prioritizes reducing the total execution time, and a lower value focuses on minimizing the Lyapunov drift.

Given the computation offloading and computing resource decisions, along with the task arrival rate at time slot $t$, the drift-plus-penalty inequality, derived from $V \geq 0$, is 
\begin{equation}
\begin{split}
&\Delta(Q(t)) + V\mathbb{E}[\Phi (t)|Q(t)] \leq D + V\mathbb{E}[\Phi (t)|Q(t)]\\
&+\mathbb{E} \bigg\{ \sum_{i=1}^{I} Q_i^{m}(t) \bigg( A_i(t) - \sum_{j=1}^J \tau R_{i,j}^{m2u}(t) \bigg)  \bigg|Q(t) \bigg\}\\
&+\mathbb{E}\bigg\{\sum_{j = 1}^J Q_j^{u}(t) \bigg( \tau R_{i,j}^{m2u}(t)  \\
&\quad\quad\quad\quad\quad\ \;- \bigg(\sum_{k = 1}^K \tau R_{j,k}^{u2v}(t) - \tau f_{i,j}^u(t)\bigg)\bigg)  \bigg| Q(t) \bigg\}\\
&+\mathbb{E} \bigg\{ \sum_{k=1}^{K} Q_k^{v}(t)\bigg(\sum_{j = 1}^J \tau R_{j,k}^{u2v}(t)-\tau f_{j,k}^v(t)\bigg)\bigg|Q(t) \bigg\},
\end{split}
\end{equation}
in which
\begin{equation}
\begin{aligned}
D &= \frac{1}{2}\bigg\{\sum_{i=1}^{I}\big[\tau R_j^{umax} + (A_i^{max})^2\big] + \sum_{j=1}^{J}\big[(\tau R_k^{vmax} -  \\
& \tau f_j^{max})^2+ (\tau R_j^{umax})^2\big] + \sum_{k=1}^{K} \big[(\tau R_v^{umax})^2 - (\tau f_k^{vmax} )^2 \big] \bigg\}.
\end{aligned}
\end{equation}
The detailed proof is in Appendix A.

Based on the above discussion, the stochastic optimization $\mathscr{P}0$ is transformed to focus on minimizing the Lyapunov drift-plus-penalty upper bound in each time slot $t$.	
Consequently, by disregarding the fixed term $D$, $\mathscr{P}0$ is reformulated into a deterministic $\mathscr{P}1$, which is applicable to any time slot, i.e.,

\begin{equation}
\begin{split}
&\mathscr{P}1:\min_{\mathcal{O},\mathcal{S},\mathcal{F}^u,\mathcal{F}^v} \mathcal{C}(\mathcal{O},\mathcal{S},\mathcal{F}^u,\mathcal{F}^v) = \\
&V\Phi(t)(\mathcal{O},\mathcal{S},\mathcal{F}^u,\mathcal{F}^v) - \sum_{i=1}^{I} Q_i^{m}(t)\bigg(\sum_{j=1}^J \tau  R_{i,j}^{m2u}(t) \bigg)\\
&\qquad +\sum_{j = 1}^J Q_j^{u}(t) \bigg( \tau R_{i,j}^{m2u}(t) - \bigg(\sum_{k = 1}^K \tau  R_{j,k}^{u2v}(t) \\
&- \tau f_{i,j}^u(t)\bigg)\bigg) +\sum_{k=1}^{K} Q_k^{v}(t)\bigg(\sum_{j = 1}^J \tau  R_{j,k}^{u2v}(t)-\tau f_{j,k}^v(t)\bigg),\\
&\text{s.t. } (\ref{c2})-(\ref{c5}), (\ref{c7}), (\ref{c9}). \notag\\ 
\end{split}
\end{equation}
Therefore, the solution of $\mathscr{P}1$ depends on the current information.
However, $\mathscr{P}1$ presents a complex MIP and non-convex optimization problem, incorporating binary variables $\mathcal{O}$ and $\mathcal{S}$, as well as continuous variables $\mathcal{F}^u$ and $\mathcal{F}^v$.	
The non-convex objective function complicates deriving an optimal solution, especially due to the interplay among variables such as computation offloading and resource allocation in dynamic networks 	Hence, we design a HASAC scheme to effectively tackle this complexity.

\begin{figure*}[!t]
\centering
\includegraphics[width=18cm]{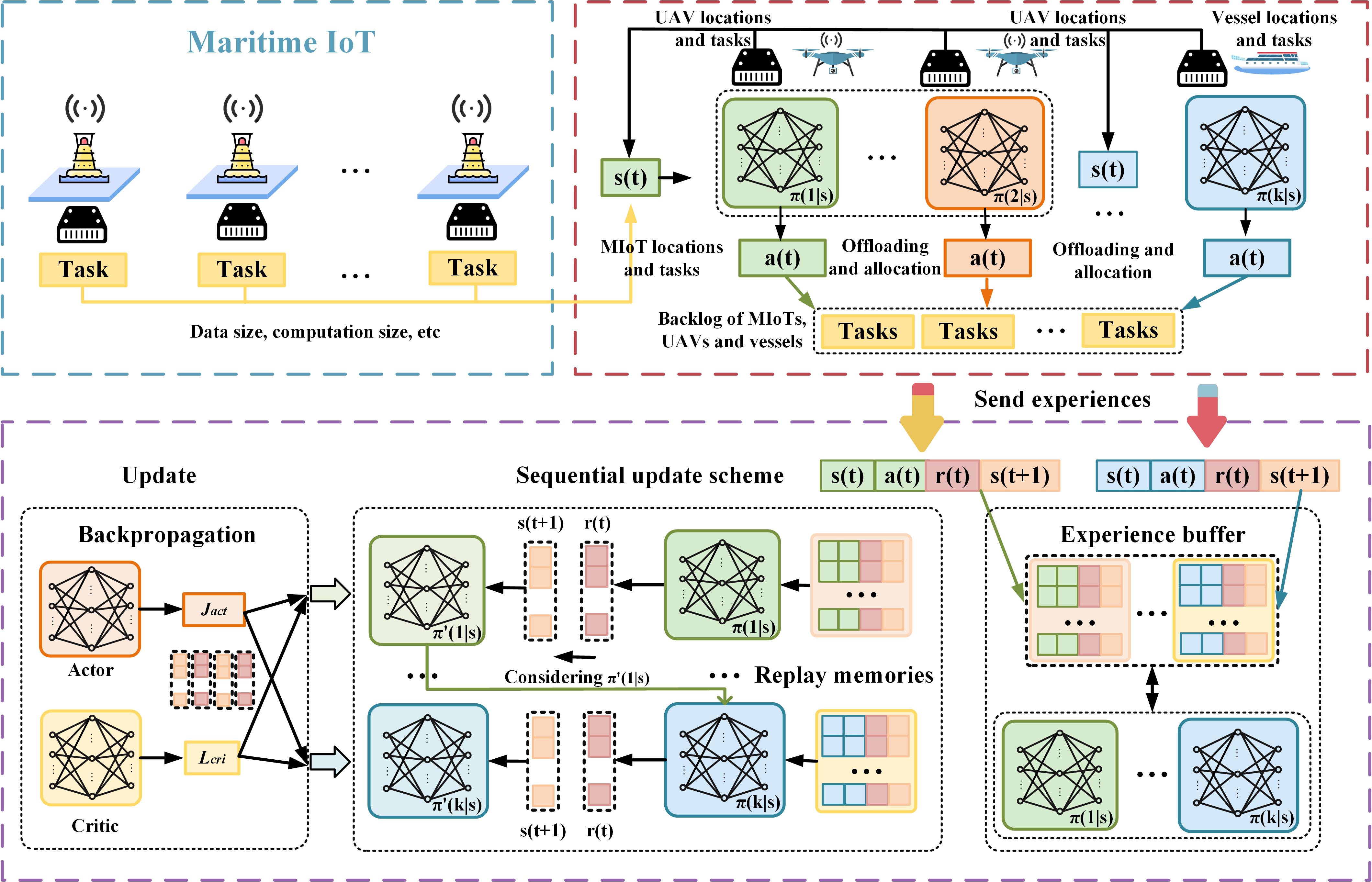}
\caption{The HASAC framework in uncertain maritime scenario.}
\label{f3}
\end{figure*}

\section{Algorithm Design}\label{s5}

\subsection{MG Framework} 
We optimize the computation offloading and resource allocation for MIoTs, UAVs, and vessels, to minimize the total execution time.	
Thus, the problem is formulated as a partially observable MG with $J$ UAV agents and $K$ vessel agents, totaling $J+K$ agents.	
In particular, the interaction process for computation offloading and resource allocation is characterized by the tuple $\langle \mathcal{S}, \{\mathcal{O}_j^u, \mathcal{O}_k^v\}_{j \in J, k \in K}, \{\mathcal{A}_j^u, \mathcal{A}_k^v\}_{j \in J, k \in K}, \mathcal{R}, \gamma \rangle$. 
Here, $\mathcal{S}$ represents the potential environmental states, including location and queue information for MIoTs, UAVs, and vessels, as well as current task infromation.	
$\mathcal{O}_j^u$ and $\mathcal{O}_k^v$ define the observation spaces for UAV agent $j$ and vessel agent $k$, respectively, with each agent observations constituting a subset of the environmental state $s(t) \in \mathcal{S}$. 
$\mathcal{A}_j^u$ and $\mathcal{A}_k^v$ represent the action sets available to UAV agent $j$ and vessel agent $k$.
At each state $s(t) \in \mathcal{S}$, UAV agent $j$ and vessel agent $k$ implement policies $\pi_j^u : S \rightarrow \mathcal{A}_j^u$ and $\pi_k^v : S \rightarrow \mathcal{A}_k^v$.	
$\mathcal{R}$ denotes the reward function, and $\gamma$ represents the discount factor.
In the following, we detail the environmental state, observation space, action space, and reward function for UAVs and vessels for each time slot $t$ \cite{MARL_BAI_2023}.

\subsubsection{Environment State Space}
For each time slot $t$, the environmental state is denoted as $s(t) \in \mathcal{S}$.
Apart from queue and current task arrival information, $s(t)$ encompasses positional data of MIoTs, UAVs, and vessels.
Thus, the state space $s(t)$ is represented as	
\begin{equation}
\begin{split}
s(t) = \left\{L^m(t),L^u(t),L^v(t), Q^m(t), Q^u(t), Q^v(t)\right\},
\end{split}
\end{equation}
where $L^m(t)$, $L^u(t)$, $L^v(t)$, $Q^m(t), Q^u(t)$, and $Q^v(t)$ denote, respectively, the location sets and backlog queues of MIoT devices, UAVs, and vessels at time slot \(t\).

\subsubsection{Observation Space}
In the partially observable MEC environment, local observations for UAV agent $j$ and vessel agent $k$ at time slot $t$ are detailed below.
\paragraph{UAV Agent Local Observation Space}
For each time slot $t$, UAV agent $j$ observes MIoTs and UAVs states, i.e.,
\begin{equation}
\mathcal{O}_j^u(t) = \{L^m(t),L^u(t), Q^m(t), Q^u(t)\}.
\end{equation}
\paragraph{Vessel Agent Local Observation Space}
At each time slot $t$, vessel agent $k$ observes UAVs and vessels states, i.e.,
\begin{equation}
\mathcal{O}_k^v(t) = \{L^u(t),L^v(t), Q^u(t), Q^v(t)\}.
\end{equation}

\subsubsection{Action Space}
In accordance with problem $\mathscr{P}1$ and the associated observational data, each UAV and vessel make decisions of action from action spaces.
For each time slot $t$, the action space of UAV agent $j$ is $\mathcal{A}_j^u = \{\mathcal{O}, \mathcal{F}^u\}$, where $\mathcal{O}$ indicates UAV computation offloading decisions, and $\mathcal{F}^u$ denotes UAV resource allocation.	
At each time slot $t$, the action space of vessel agent $k$ is $\mathcal{A}_k^v = \{\mathcal{S},\mathcal{F}^v\}$, where $\mathcal{S}$ represents vessel computation offloading decisions, and $\mathcal{F}^v$ denotes the resource allocation set for vessels.	

\subsubsection{Reward Function}
Following the coordinated actions of both UAVs and vessels, the environment provides a reward $r(t)$ to assess the effectiveness of joint actions. In accordance with the objectives in problem $\mathscr{P}1$, the primary goal of the agents is to minimize the total execution time, which is formulated as $\mathcal{C}(\mathcal{O},\mathcal{S},\mathcal{F}^u,\mathcal{F}^v)$. Therefore, the current system reward function can be defined as
\begin{equation}
r(t) = \mathcal{C}(\mathcal{O},\mathcal{S},\mathcal{F}^u,\mathcal{F}^v).
\end{equation}

\subsection{HASAC-based Solution}
The distinctive observation and action spaces for UAV and vessel agents often lead to instabilities in training and challenges in convergence \cite{HARL_2024,NEURIPS2021_Advances}.
In Fig. \ref{f3}, to handle the complexity of the MG involving $J+K$ agents, we integrate the soft actor-critic algorithm and a cooperative multi-agent framework.	
In detail, by utilizing insights from the multi-agent advantage decomposition lemma, we employ a sequential update process to enhance collaboration among the heterogeneous multi-agent systems.	
In this process, each agent updates its policy one at a time in a fixed order.  This method ensures that each agent update considers the most recent updates from the other agents, thereby preventing conflicting updates that could destabilize the learning process.
The method is triggered by partitioning the collective advantage into sequential assessments, each considering the actions of prior agents.		
Then, a collaborative reward is allocated among all $J+K$ agents to achieve the optimization objective.	

As for a set of $J+K$ agents, denoted by $i_{1:n}$, functioning within state $s$ and undertaking actions $\boldsymbol{a}^{i_{1: n}}$, the multi-agent soft Q-function $Q_{\boldsymbol{\pi}}^{i_{1: n}}(s, \boldsymbol{a}^{i_{1: n}})$ is defined, with $-i_{1:n}$ representing the complementary set of agents.	
This function quantifies expected cumulative discounted future rewards, denoting anticipated returns within the present state-action framework.	
Hence, the multi-agent soft Q-function is modeled as
\begin{equation}
\begin{split}
Q_{\boldsymbol{\pi}}^{i_{1: n}}(s, \boldsymbol{a}^{i_{1: n}}) &\triangleq \mathbb{E}_{\mathbf{a}^{-i_{1: n}} \sim \boldsymbol{\pi}^{-i_{1: n}}}\Big[Q_{\boldsymbol{\pi}}(s, \boldsymbol{a}^{i_{1: n}}, \mathbf{a}^{-i_{1: n}}) \\
&\qquad \qquad \qquad   +\alpha \sum_{i \in -i_{1: n}} \mathcal{H}(\pi^i(\cdot \vert s))\Big],
\end{split}\label{e39}
\end{equation}
where $\mathbb{E}_{\mathbf{a}^{-i_{1: n}} \sim \boldsymbol{\pi}^{-i_{1: n}}}$ represents the expected actions from agents outside $i_{1:n}$, according to the strategy $\boldsymbol{\pi}^{-i_{1: n}}$. 
The temperature parameter $\alpha$ adjusts the impact of entropy regularization during the optimization of the policy.	
$\sum_{i \in -i_{1: n}} \mathcal{H}(\pi^i(\cdot | s))$ represents the sum of entropies of actions taken by all agents excluding $i_{1:n}$ in state $s$.

The algorithm framework aims to design optimal strategies for computation offloading and resource allocation with minimized possible time cost.
In detail, the long task completion time indicates suboptimal results, underscoring the current strategy limitations and the imperative for improvement.	
Furthermore, HASAC focuses on learning strategy parameters through the minimization of expected Kullback-Leibler (KL) divergence. Consequently, the joint policy is presented as	
\begin{equation}
\begin{split}
\boldsymbol{\pi}_{new}=\arg \min _{\boldsymbol{\pi^{\prime}} \in \boldsymbol{\Pi}} \mathrm{D}_{\mathrm{KL}}\left(\boldsymbol{\pi}^{\prime}(\cdot \vert s) \| \frac{\exp \left(\frac{1}{\alpha} Q_{\boldsymbol{\pi}_{old}}(s, \cdot)\right)}{Z_{\boldsymbol{\pi}_{old}}(s)}\right),
\end{split}
\end{equation}
where $\boldsymbol{\Pi}$ is the comprehensive set of feasible policies accessible to an agent.	
$\mathrm{D}_{\mathrm{KL}}$ represents the KL divergence, an information projection that can conveniently map the improved policy onto the desired set of policies.
$\boldsymbol{\pi}^{\prime}(\cdot \vert s)$ denotes the new policy probability distribution over actions, given the current state $s$.	
$\exp(\cdot)$ is the exponential function.
$Z(\cdot)$ indicates to normalize the distribution, to adjust the exponential output of the soft Q-function to validate the target probability distribution.

The function approximators, specifically deep neural networks (DNNs), are utilized to model both the centralized soft Q-function $Q_\theta (s_t, \mathbf{a}_t)$, and the decentralized policies $\pi^{i_n}_{{\phi}^{i_n}}$ for each agent $i_m$ are parameterized by $\theta$ and ${\phi}^{i_n}$, respectively.	
The optimization of these networks proceeds via iterative application of stochastic gradient descent, focusing on minimizing the bellman residual for the Q-function, i.e.,
\begin{equation}
\begin{split}
&J_Q(\theta)=\mathbb{E}_{(\mathrm{s}_t, \mathbf{a}_t) \sim \mathcal{D}}\bigg[\frac{1}{2}(Q_\theta(\mathrm{s}_t, \mathbf{a}_t)-(r(\mathrm{s}_t, \mathbf{a}_t)\\
&\qquad\qquad\qquad\qquad\qquad+\gamma \mathbb{E}_{\mathrm{s}_{t+1} \sim P}[V_{\bar{\theta}}(\mathrm{s}_{t+1})]))^2\bigg].
\end{split}
\end{equation}
Here, $\mathcal{D}$ represents the replay buffer, utilized for storing historical experiences.
The discount factor $\gamma$ measures the significance of future rewards, and $V_{\bar{\theta}}(\mathrm{s}_{t+1})$ estimates the value function of the subsequent state $\mathrm{s}_{t+1}$, parameterized by $\bar{\theta}$. 
Specifically, $\bar{\theta}$ denotes the parameters of the target network, which is a delayed replica of the Q-network intended to stabilize the training process.
Similarly, the parameters of the policy $\pi^{i_n}_{{\phi}^{i_n}}$ are optimized by minimizing the expected KL divergence, which is expressed as
\begin{equation}
\begin{split}
&J_{\pi_{i_n}}(\phi^{i_n}) = \mathbb{E}_{s_t \sim \mathcal{D}} \bigg[ 
\mathbb{E}_{a_t^{i:1:n-1} \sim \pi^{i:1:n-1}_{\phi^{i:1:n-1}_{\text{new}}}, a_t^{i_n} \sim \pi^{i_n}_{\phi^{i_n}}}\\ 
&\qquad  \Big[ \alpha \log \pi^{i_n}_{\phi^{i_n}}(a_t^{i_n} | s_t) - Q^{i:1:n}_{\pi_{\text{old}};\theta}(s_t, a_t^{i:1:n-1}, a_t^{i_n})\Big]
\bigg].
\end{split}
\end{equation}

\begin{algorithm}[t]
\caption{Heterogeneous-Agent Soft Actor-Critic}
\label{a1}
\begin{algorithmic}[1]
\REQUIRE Temperature parameter $\epsilon$, Polyak coefficient $\iota$, batch size $B$, number of agents $n$, episodes $K$, steps per episode $J$.
\STATE \textbf{Initialize:} Critic networks $\phi_1$ and $\phi_2$, policy networks $\{\theta_i\}_{i \in \mathcal{N}}$, replay buffer $\mathcal{D}$, target parameters $\phi_{\text{targ},1} \leftarrow \phi_1$, $\phi_{\text{targ},2} \leftarrow \phi_2$.
\FOR{$k = 0$ to $K-1$}
    \FOR{agent $i \in \mathcal{N}$, episode $k$, step $j$}
        \STATE Observe state $o_i^j$ and select action $a_i^j \sim \pi_{\theta_i}(o_i^j)$.
        \STATE Execute action $a_i^j$ in the environment.
        \STATE Observe next state $o_{i,j+1}$ and reward $r_i^j$.
        \STATE Store $(o_{i,j}, a_{i}^j, r_{i}^j, o_{i,j+1})$ into $\mathcal{D}$.
        \STATE Sample a random batch of samples from $\mathcal{D}$.
        \STATE Compute the critic targets using (\ref{e44}).
    \ENDFOR
    \STATE Update Q-functions by one step of gradient descent leveraging (\ref{e43}).
    \STATE Randomly select a permutation of agents $i_{1:n}$.
    \FOR{agent $i_m = i_1, \ldots, i_n$}
        \STATE Update agent $i_m$ by using (\ref{e45}).
    \ENDFOR
    \STATE Smoothly update the target critic network utilizing (\ref{e46}).
\ENDFOR
\end{algorithmic}
\end{algorithm}

\subsection{HASAC Algorithm Implementation}
As depicted in Algorithm \ref{a1}, the centralized agent in each network includes the main network, target network, policy gradient, loss function, and replay memory.
we utilize two soft Q-functions to mitigate positive bias in the policy improvement step and counteract overestimation. 	
The actor manages learning the computation offloading and resource allocation strategies and generates actions based on the state via a fully connected network.	
Two evaluators are employed to assess the actor computation offloading and resource allocation strategy, to enhance the training efficiency.	
Similarly, the critic network is structured as a fully connected neural network, with neurons in both the actor and critic representing the inputs of states and actions, respectively.	
The replay memory stores historical experience information that is used to update the actor and critic, including actions, states, rewards, and subsequent states.	

To address the computational challenges associated with the HASAC algorithm, we separate the problem into two phases: centralized training and distributed execution \cite{Decentralized_2024,QoE_2022,Collaborative_2024}. During the training phase, the algorithm is trained centrally, allowing for intensive computation using high-performance computing resources. This phase can be performed offline, and the learned policies are stored for execution. In the execution phase, the algorithm is decentralized, with each UAV and vessel independently implementing the trained policies. This distribution of execution minimizes the computational load on individual agents and ensures that the system can scale efficiently in real-time operations.

During the training process, the agent initially collects local data (line 1). Specifically, each local state is transmitted during each training segment. 
Subsequently, the agent actor generates joint offloading and resource allocation actions based on the local state (line 4).	
Hence, the networked environment generates rewards based on the current action and state (lines 5-6).	
The tuples of actions, states, rewards, and subsequent states are stored in the replay memory $\mathcal{D}$ (line 7). 
Then, a batch of tuples $B$ is randomly sampled from the replay memory $\mathcal{D}$ to update the network (line 8). The joint actions and local state are inputted into the critic main network for evaluation.	
Then, the critic target network is then updated based on its main network (line 9). 
In HASAC, the mean squared error serves as the loss function to train the Q critic network, utilizing the optimal Bellman equation (lines 11-12), given by
\begin{equation}
\begin{split}
\phi_i = \arg\min_{\phi_i} \frac{1}{B} \sum_t (y_t - Q_{\phi_i}(s_t, a_t))^2,
\end{split}\label{e43}
\end{equation}
in which
\begin{equation}
\begin{split}
&y_t = r + \gamma \Big(\min_{i=1,2} Q_{\phi_{\text{targ},i}}(s_{t+1},a_{t+1}) \\
&\quad - \alpha \sum_{i=1}^{n} \log \pi_{\theta_i}(a_{i,t+1}|o_{i,t+1}) \Big), \boldsymbol{a}_{t+1} \sim \boldsymbol{\pi_{\theta}}(\cdot|s_{t+1}).
\end{split}\label{e44}
\end{equation}

Then, the actor network is responsible for mapping the states to the actions. Hence,  Its main task is to obtain the optimal policy by minimizing the KL divergence (line 14), i.e.,
\begin{equation}
\begin{split}
&\theta_{i_m}^{\text{new}} = \arg\max_{\theta_{i_m}} \frac{1}{B} \sum_t \Big[ \min_{i=1,2} Q_{\phi_i} \Big( s_t, a_{i_m}^{t-1}, \big(a_{i_{1:m-1}}^{t}\big), \\
&\qquad\qquad\qquad a_{i_m}^{\theta_{i_m}}, a_{i_{m+1:n}}^{t} \Big) - \alpha \log \pi_{\theta_{i_m}}^{\iota} \big( a_{i_m}^{\theta_{i_m}} | o_{i_m}^t \big) \Big].
\end{split}\label{e45}
\end{equation}
Finally, the target critic network is updated smoothly (line 16):
\begin{equation}
\begin{split}
\phi_{\text{targ},i} \leftarrow \rho \phi_{\text{targ},i} + (1 - \rho) \phi_i.
\end{split}\label{e46}
\end{equation}

\begin{table}[!t]
\caption{Simulation Parameters}
\centering
\begin{tabular}{|m{1.8cm}<{\centering}|m{1.8cm}<{\centering}||m{1.8cm}<{\centering}|m{1.8cm}<{\centering}|}
\hline
\textbf{Parameter} & \textbf{Value} & \textbf{Parameter} & \textbf{Value} \\ \hline
$\zeta_L$ & $2.3$ & $\zeta_{NL}$ & $34$ \\ \hline
$\alpha$ & $5.0188$ & $\beta$ & $0.3511$ \\ \hline
$h_i^m$, $h_j^u$ & 0 m, 30 m & $\varphi_c$ & 2 GHz \\ \hline
$C_0$ & $3\times10^8$ m/s & $B_0$ & 1 MHz \\ \hline
$P_i^m$ & 0.5 W & $N_G$ & -114 dBm \\ \hline
$G_0$ & -50 dB &$L$& 2 \\ \hline
$B$  & 20 MHz & $P_{j,k}^u$ & 5 W \\ \hline
$\lambda_i$ & 15 & $c_i(t)$ & 270 cycles/bit \\ \hline
$f_j^{umax}$ & $10^9$ cycles/s & $f_k^{vmax}$ & $10^{10}$ cycles/s \\ \hline 
Learning rate & $5\times 10^{-4}$ & Batch size & 1024 \\ \hline
Buffer size&  $1 \times 10^6$ & Discount factor & 0.99 \\ \hline
Temperature $\epsilon$ & 0.001 & Hidden sizes & [512, 512] \\ \hline
\end{tabular}
\label{T2}
\end{table}

\begin{figure*}[htbp]
\centering
\subfloat[Test reward under different learning rates.]{
\includegraphics[width=5.8cm]{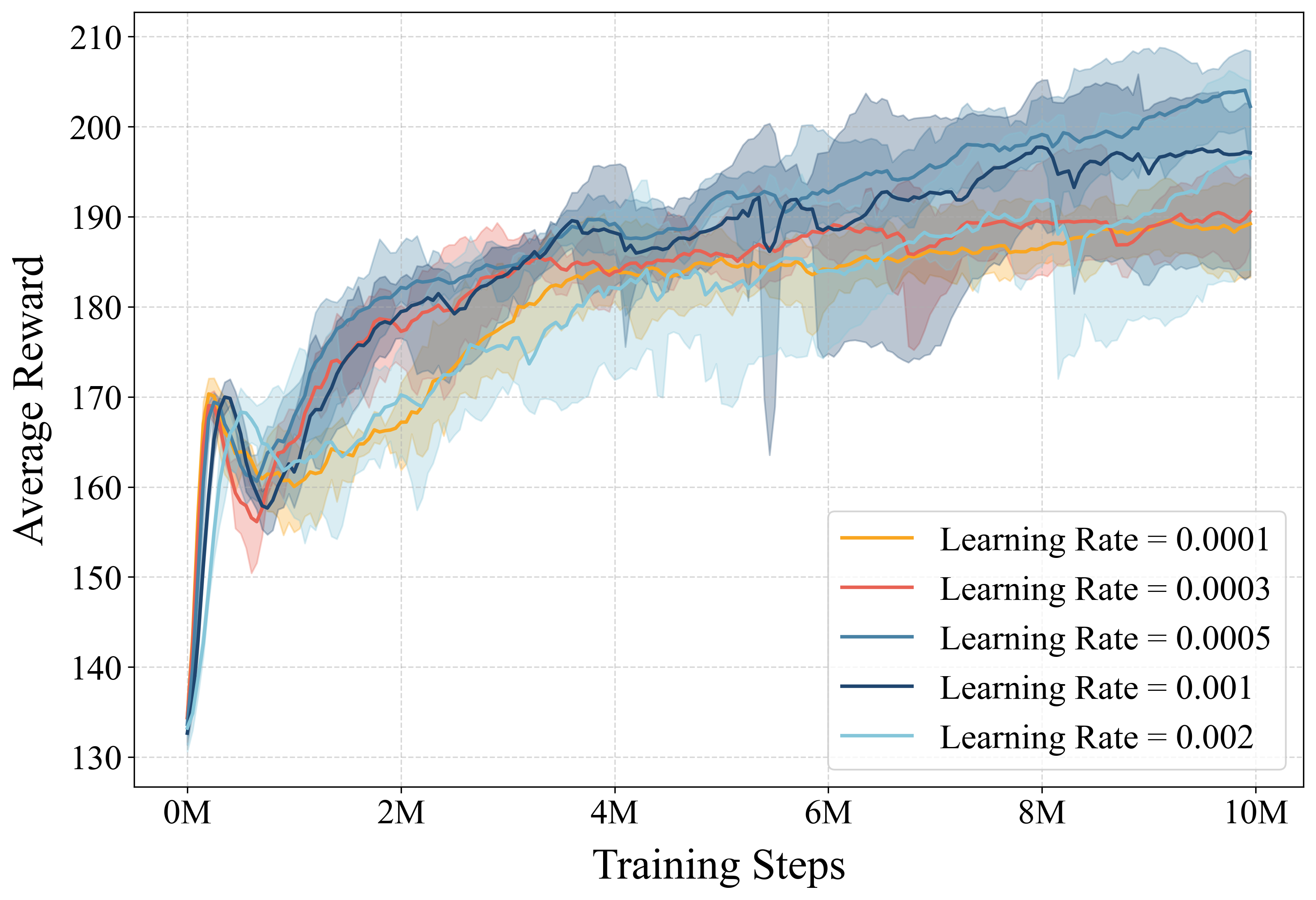}}
\hfill
\subfloat[Test reward under different hidden sizes.]{
\includegraphics[width=5.8cm]{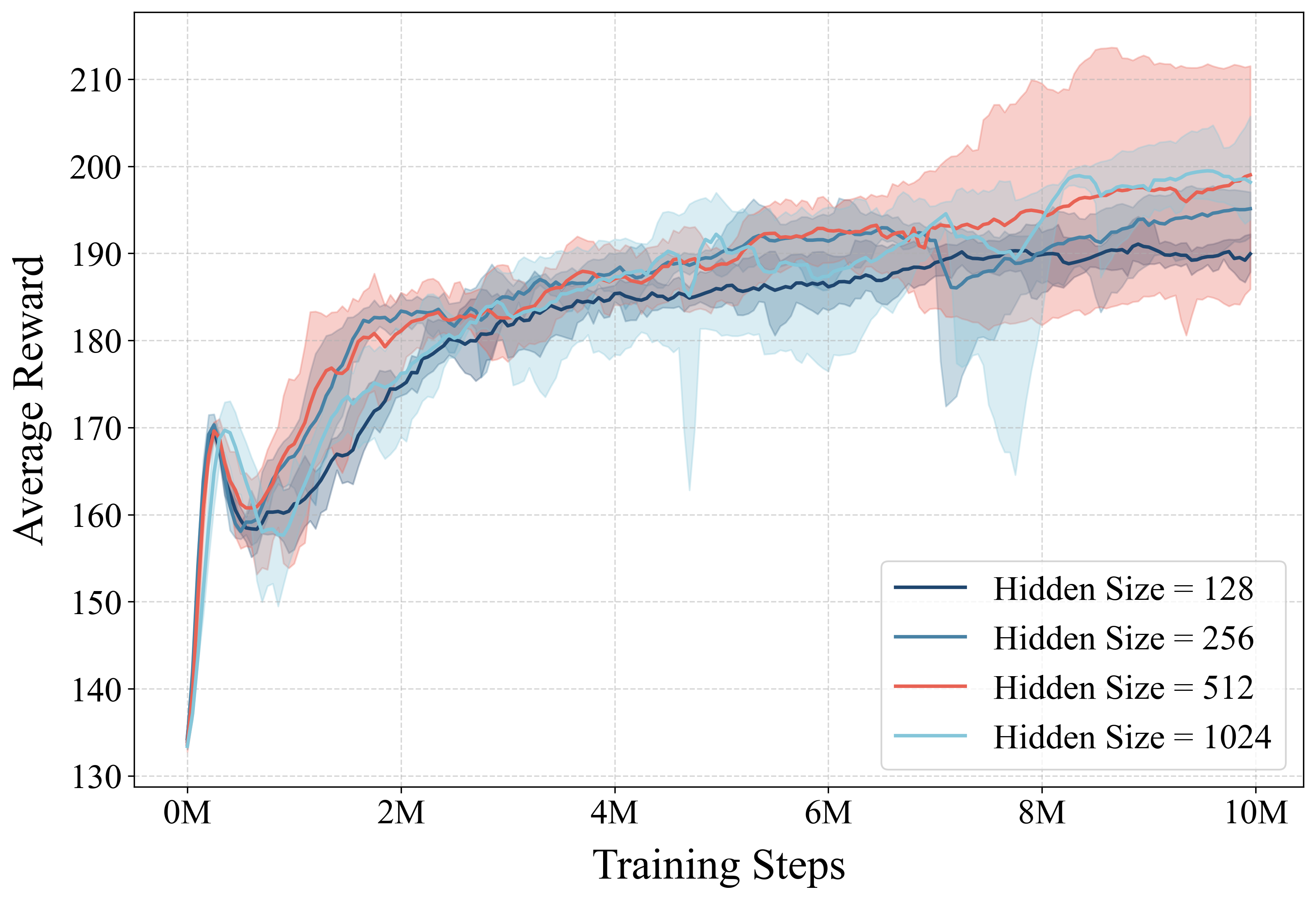}}
\hfill
\subfloat[Test reward under different activation functions.]{
\includegraphics[width=5.8cm]{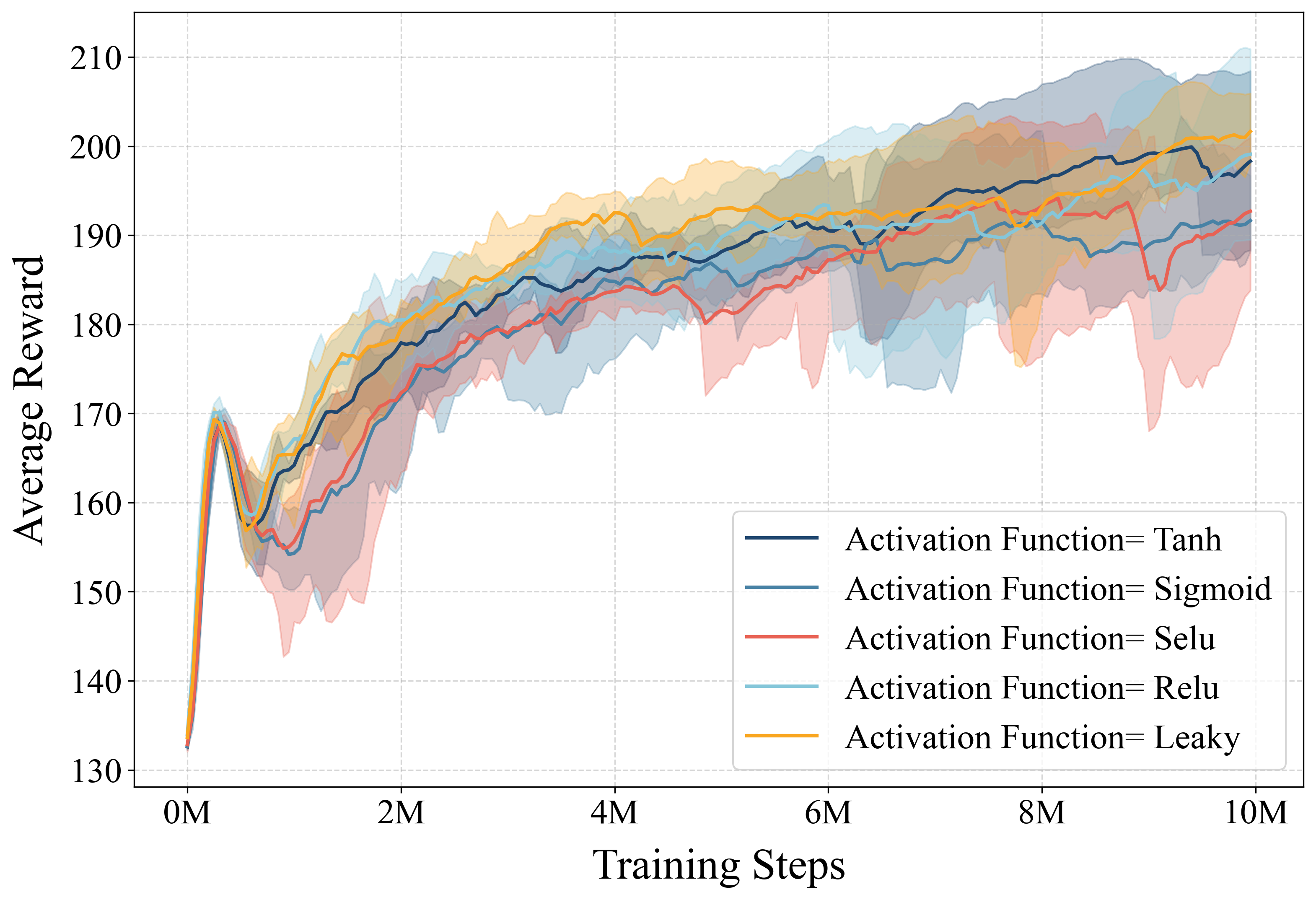}}
\caption{Impact of hyperparameter settings on the HASAC model performance: (a) Learning rate, (b) Hidden layer size, and (c) Activation function.}
\label{f4}
\end{figure*}

\section{Simulation Results}\label{s6}
In this section, we evaluate the performance of the HASAC-based computation offloading and resource allocation algorithm through extensive simulations. The algorithm is compared with the following benchmarks:

\subsubsection{Heterogeneous-Agent Advantage Actor-Critic (HAA2C)}
The HAA2C algorithm \cite{HARL_2024} extends the advantage actor-critic framework to address decision-making challenges in heterogeneous multi-agent environments.

\subsubsection{Proximity Heuristic (PH)}
This algorithm assigns tasks based on physical proximity to computing resources, prioritizing the nearest device while balancing transmission costs with available computing power.

\subsubsection{Greedy Completion Time (GCT)}
A greedy algorithm that selects the device with the shortest estimated completion time for each task, considering the transmission, computation time, and resource load.

\subsubsection{Centralized Load Balancer (CLB)}
A global load balancing algorithm that distributes tasks to the least loaded resource, normalizing the current load by processing power to maximize system efficiency.

\subsubsection{Random Offloading (RO)}
A baseline algorithm that randomly assigns tasks to available resources (local, UAV, or vessel) without considering task requirements and system state.

\begin{figure*}[htbp]
\centering
\subfloat[Average completion time.]{
\includegraphics[width=5.8cm]{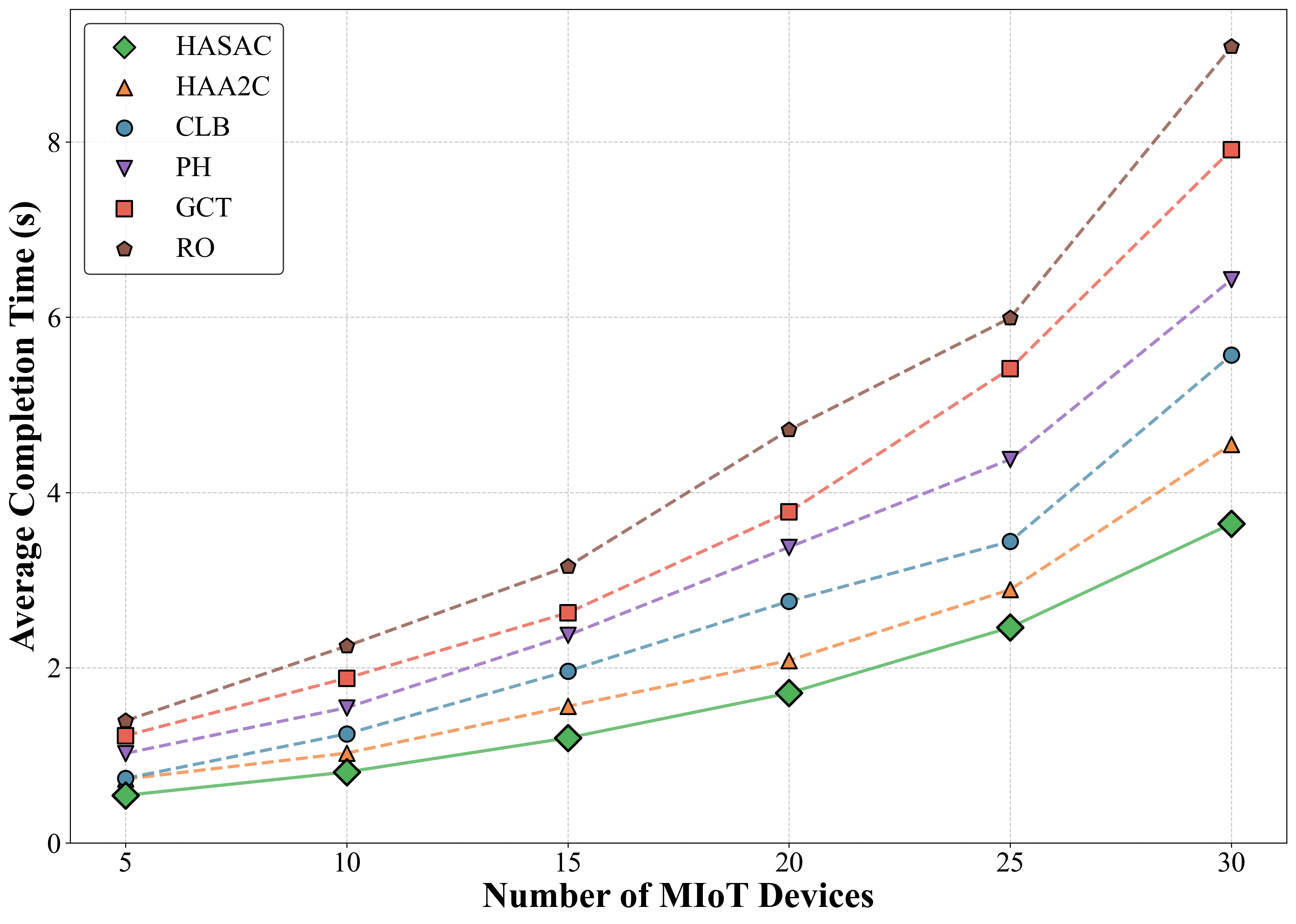}}
\hfill
\subfloat[Average response time.]{
\includegraphics[width=5.8cm]{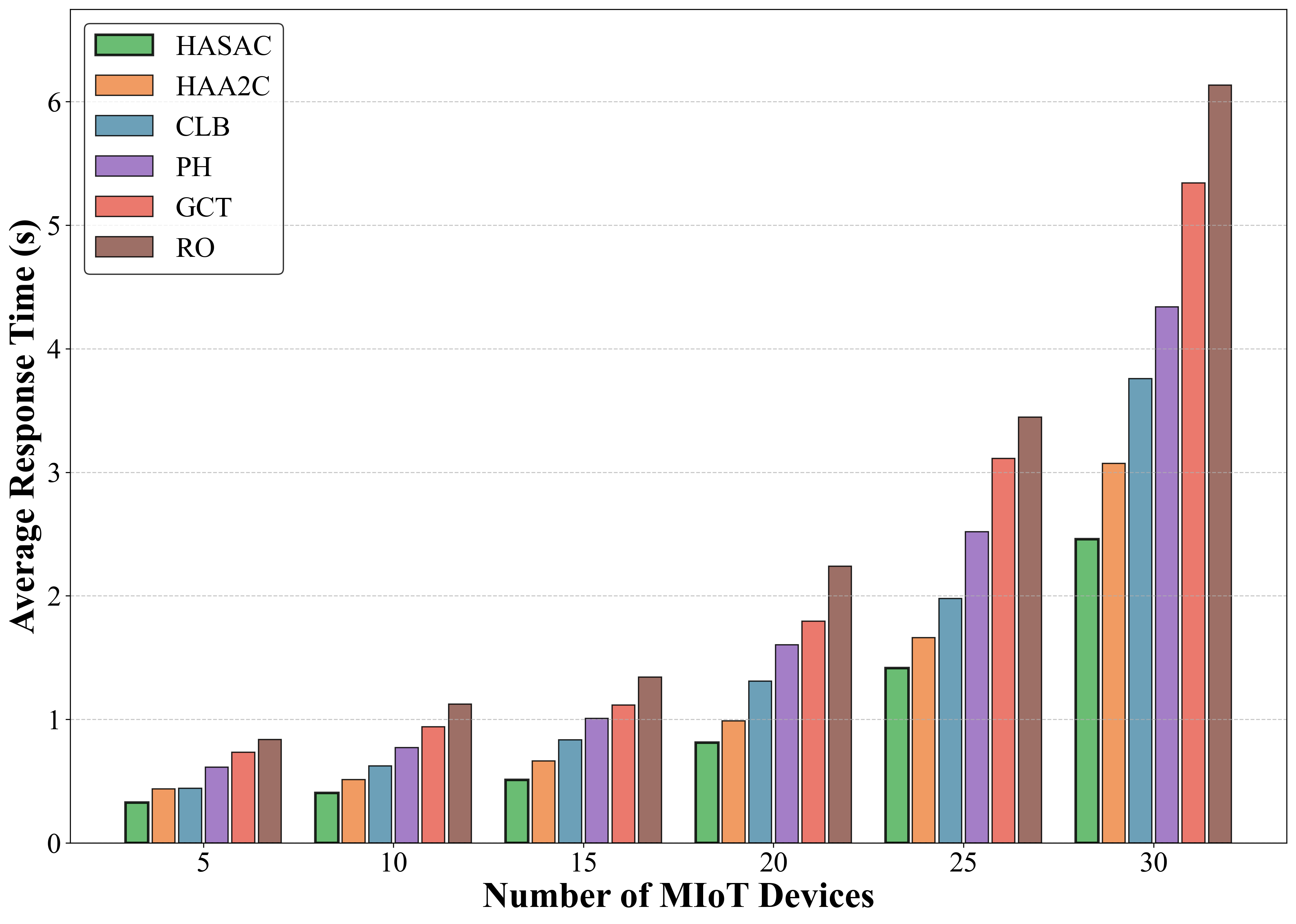}}
\hfill
\subfloat[Edge computing percentage.]{
\includegraphics[width=5.8cm]{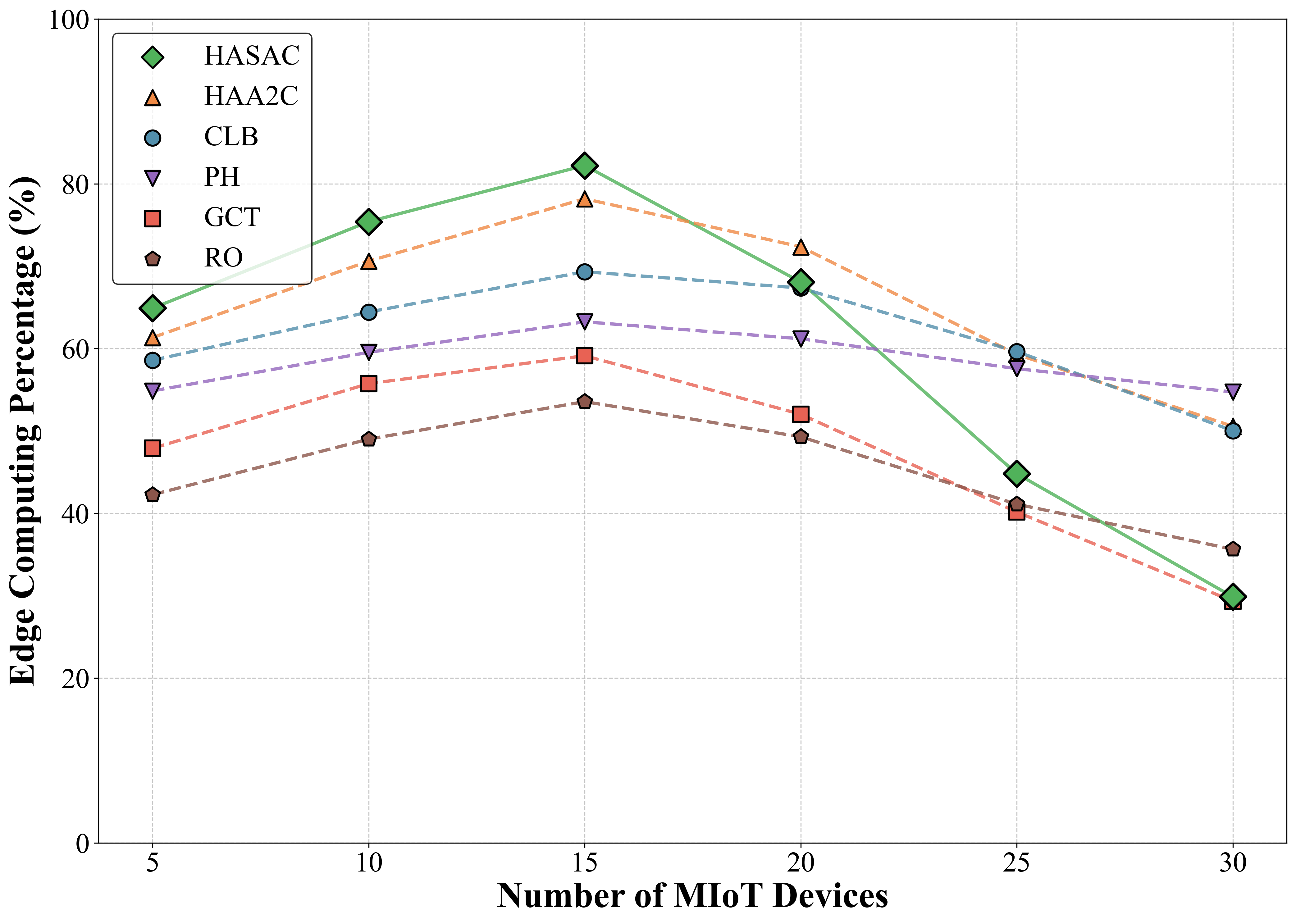}}
\caption{Performance with Different Numbers of MIoTs: (a) Average completion Time, (b) Average response time, (c) Edge computing percentage.}
\label{f5}
\end{figure*}

\begin{figure*}[htbp]
\centering
\subfloat[Average completion time.]{
\includegraphics[width=5.8cm]{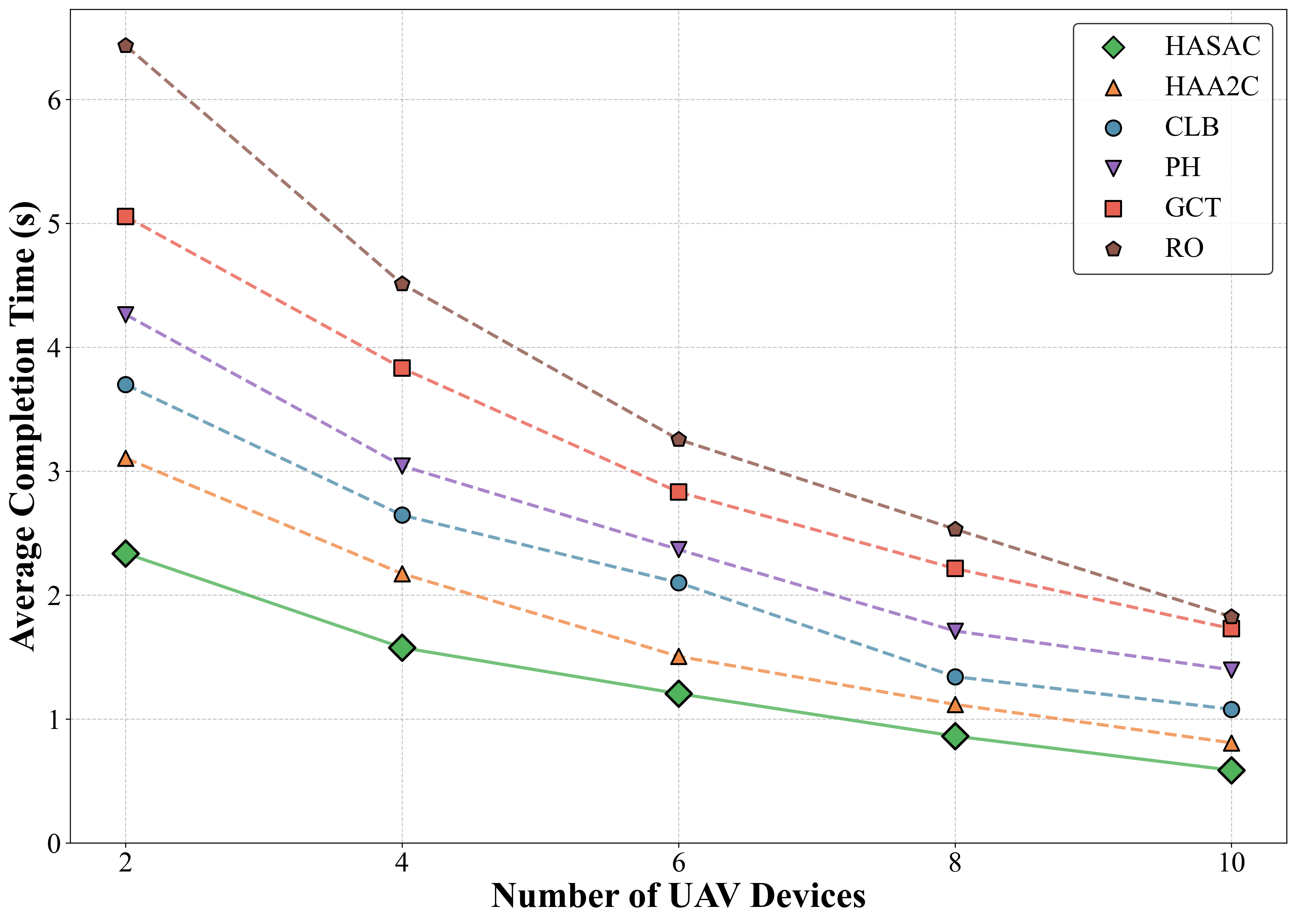}}
\hfill
\subfloat[Average response time.]{
\includegraphics[width=5.8cm]{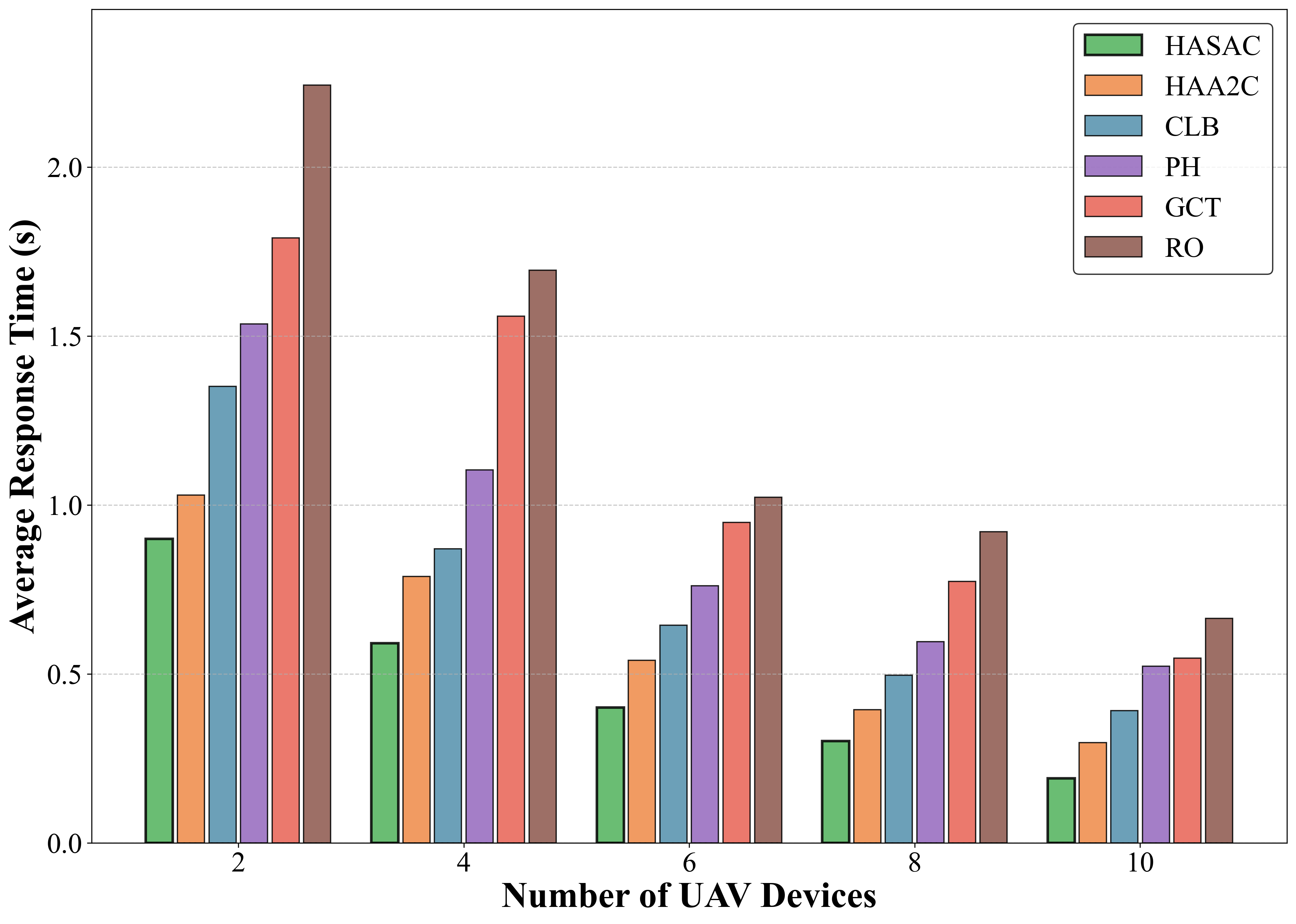}}
\hfill
\subfloat[Edge computing percentage.]{
\includegraphics[width=5.8cm]{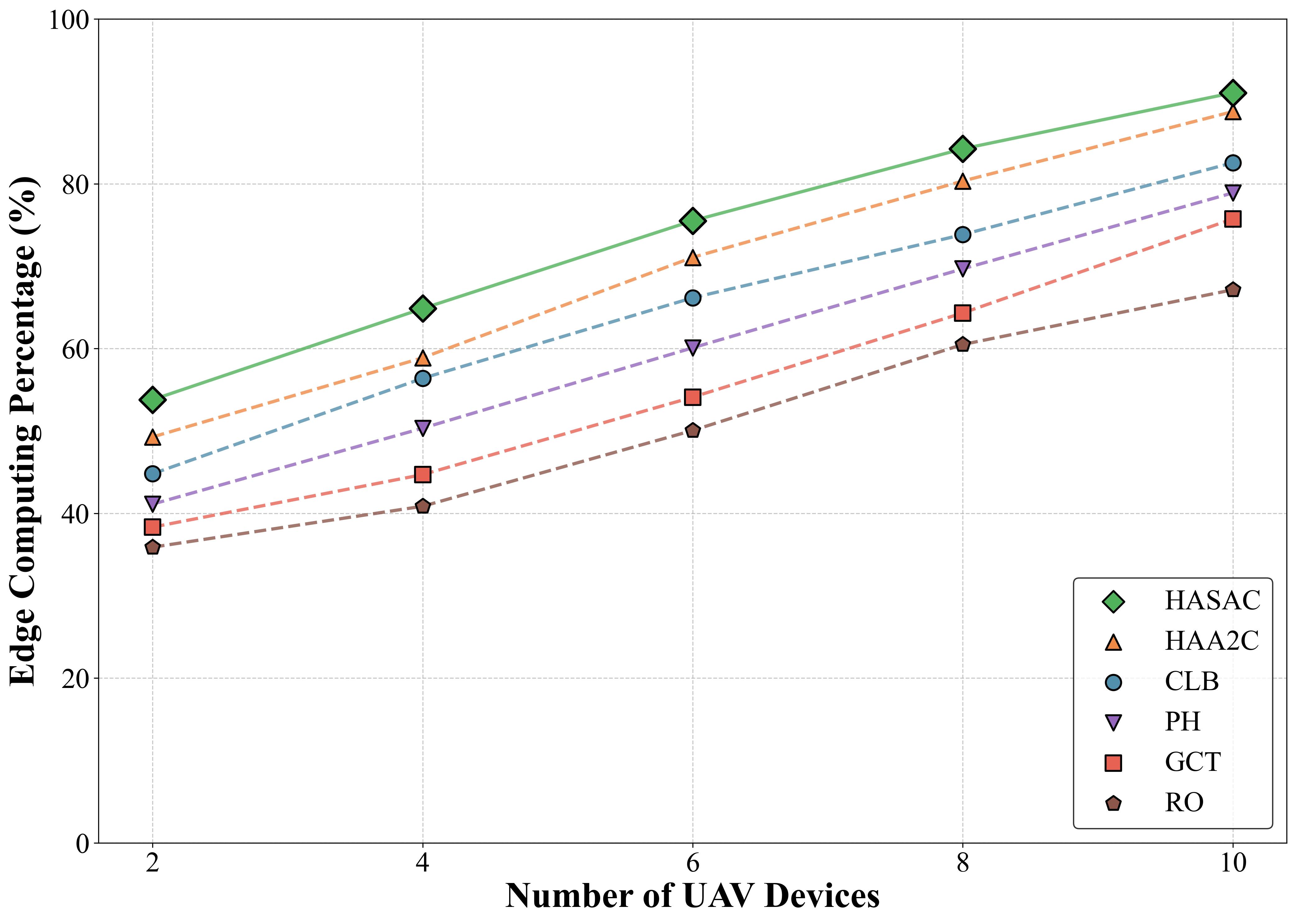}}
\caption{Performance with different numbers of UAV devices: (a) Average completion time, (b) Average response time, (c) Edge computing percentage.}
\label{f6}

\end{figure*}

\subsection{Parameter Setting}
We consider a scenario within a $1,000m\times1,000m$ area, wherein 10 MIoT devices are deployed to generate tasks, and 6 UAVs along with 2 vessels are utilized for task processing.
At each time slot, each MIoT device generates a computation-intensive task. Subsequently, the resource allocation of UAVs and vessels based on predetermined actions to process these tasks.
Concurrently, based on the offloading decisions, a predetermined number of tasks is offloaded to UAVs and vessels for execution. 
Detailed parameters are listed in Table \ref{T2} \cite{Jia-Hierarchical_2022,Computation_you_2023,Soft_2023}.
In each agent, the actor and the critics comprise an input layer, an output layer, and three hidden fully-connected layers. 
We summarize the hyperparameters of the HASAC implementation in Table \ref{T2}. 
During all simulations, we discretize the time horizon into time steps.

\subsection{Analysis of Hyperparameter Impact on Model Convergence}
In this experiment, we analyze the impact of hyperparameters such as learning rate, hidden layer size, and activation function on model convergence. By comparing the training curves and final rewards under different settings, we evaluate the specific contribution of each hyperparameter configuration to model performance.

In Fig. \ref{f4}(a), we present the training process under different learning rate settings. Larger learning rates (such as 0.001 and 0.002) accelerate early convergence but also lead to larger fluctuations. In contrast, smaller learning rates (such as 0.0001 and 0.0003) converge more slowly but maintain a more stable training process. The optimal learning rate is set to 0.0005, which achieves higher average rewards in a shorter time while maintaining good stability.

Fig. \ref{f4}(b) shows the effect of different hidden layer sizes on the model training. Smaller hidden layers (128 and 256) achieve faster training, but converge to lower rewards with greater variance. Larger hidden layers (512 and 1024) show higher final rewards, with the 512-layer network achieving a good balance between the convergence speed and final performance.

Fig. \ref{f4}(c) compares the convergence behavior under different activation functions. The LeakyReLU activation function performs best across all settings, providing the smooth and rapid convergence. In contrast, Sigmoid and SELU perform poorly, leading to slower convergence and lower final rewards. Tanh and ReLU, while relatively close in performance, still cannot outperform LeakyReLU.

In conclusion, appropriate hyperparameter selections can significantly enhance model training efficiency and convergence performance. In this experiment, the combination of a learning rate of 0.0005, a hidden layer size of 512, and the LeakyReLU activation function achieved the best results.

\subsection{Performance with Different Numbers of MIoTs}
In this section, we evaluate the performance of the proposed HASAC-based computation offloading and resource allocation algorithm against five benchmark algorithms. The evaluation focuses on three key metrics: average completion time, average response time, and edge computing percentage.

As shown in Fig.~\ref{f5}(a), the average completion time increases for all algorithms as the number of MIoT devices grows from 5 to 30. HASAC consistently achieves the lowest completion time across all configurations. When the number of MIoT devices reaches 30, HASAC maintains an average completion time of approximately 3.5 seconds, significantly outperforming GCT and RO, which reach over 8 and 9 seconds, respectively. The gap widens as the workload increases, demonstrating HASAC's superior task allocation strategy and its robustness to load escalation. CLB, PH, and HAA2C perform moderately, but their completion times increase more sharply than HASAC under heavy loads.

In Fig.~\ref{f5}(b), we observe the average response time. HASAC again achieves the lowest response time across all levels of MIoT density. Notably, while all algorithms show rising trends as the number of devices increases, HASAC's growth remains more gradual. At 30 MIoTs, HASAC's response time remains below 3 seconds, whereas RO exceeds 6 seconds and GCT reaches about 5.5 seconds. CLB and HAA2C also exhibit steady increases but stay in the mid-range (around 3-4 seconds). PH performs worse than HAA2C and CLB, indicating its limited ability to manage increased task demand.

Fig.~\ref{f5}(c) illustrates the edge computing percentage, reflecting the proportion of tasks offloaded to UAVs and vessels. HASAC demonstrates an early advantage, reaching a peak offloading rate of around 80\% at 15 MIoT devices. However, as the number of devices continues to increase, its offloading ratio gradually decreases to approximately 30\% at 30 devices, indicating resource saturation at the edge. HAA2C and CLB show more stable offloading percentages, maintaining around 60-65\%, though never surpassing HASAC's peak. PH, GCT, and RO trail significantly, with RO exhibiting the lowest utilization throughout—dropping to nearly 35\% at higher MIoT counts. This suggests that HASAC not only utilizes edge resources more aggressively but also adapts better to varying computational loads.

In summary, HASAC outperforms all benchmark algorithms across three key metrics. It demonstrates shorter completion and response times, and higher edge computing utilization—especially under increasing MIoT task density. These results confirm that HASAC is highly efficient and scalable for dynamic maritime computation offloading scenarios.

\begin{figure*}[htbp]
\centering
\subfloat[Average completion time.]{
\includegraphics[width=5.8cm]{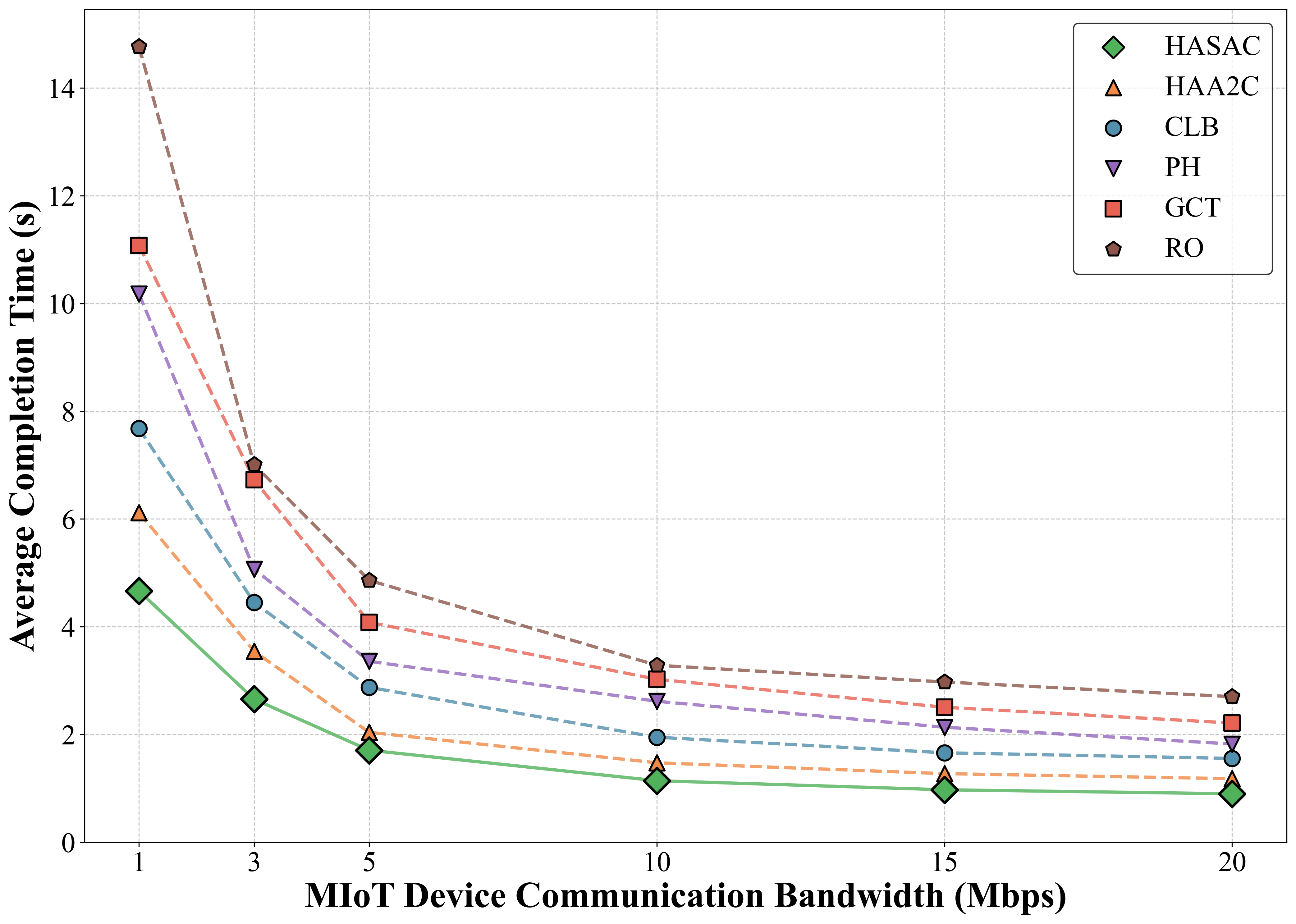}}
\hfill
\subfloat[Average response time.]{
\includegraphics[width=5.8cm]{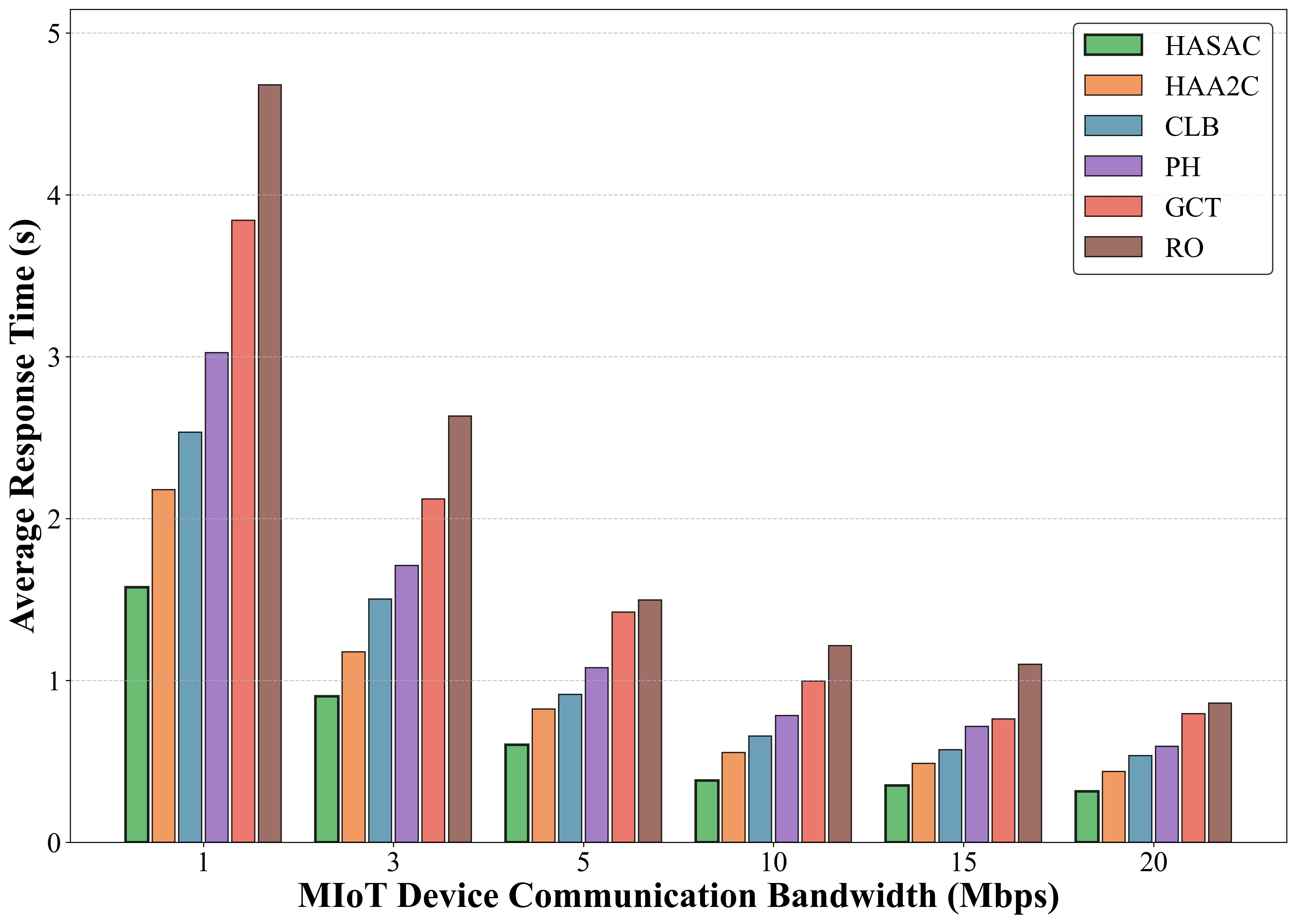}}
\hfill
\subfloat[Edge computing percentage.]{
\includegraphics[width=5.8cm]{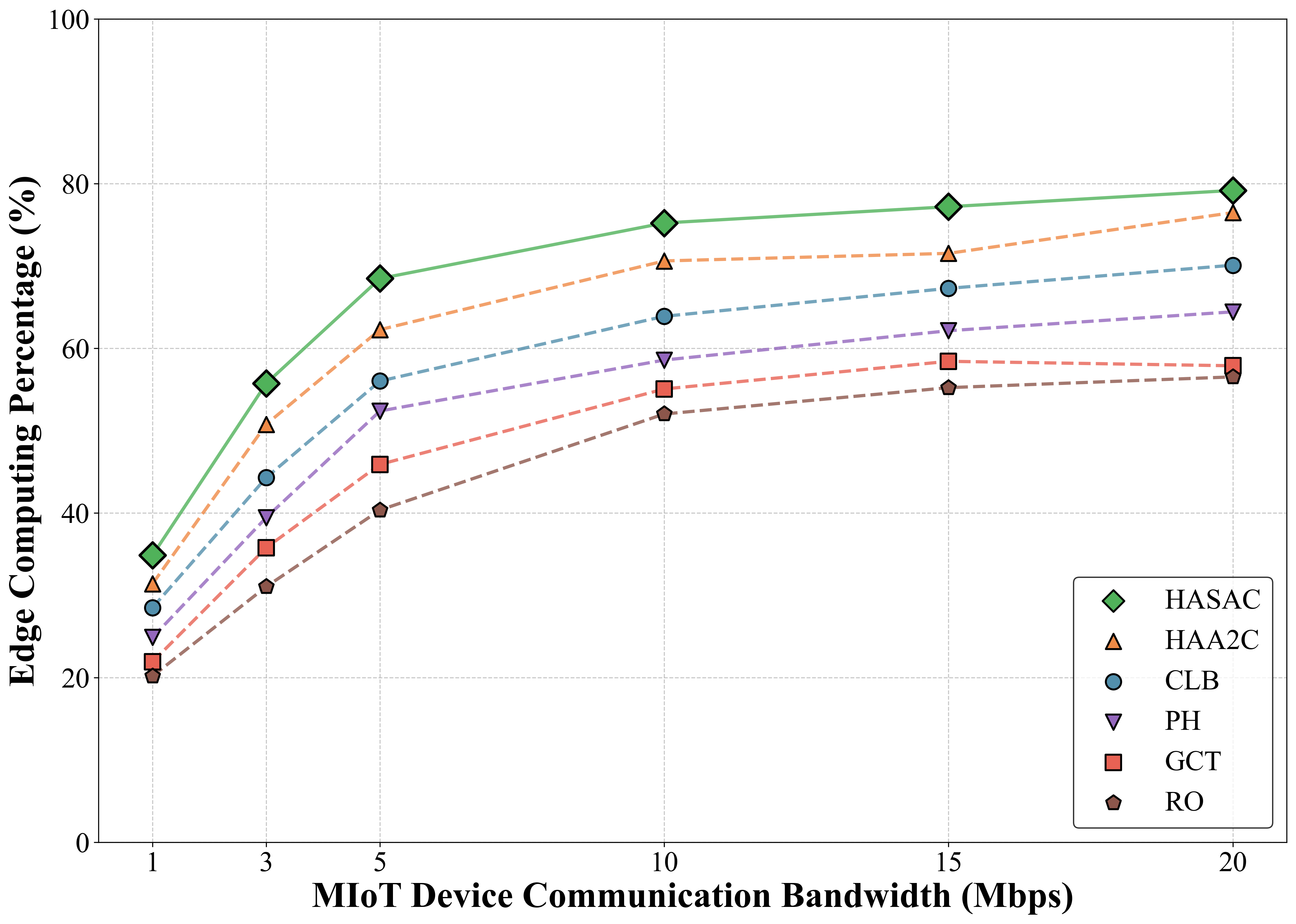}}
\caption{Performance with different MIoT device communication bandwidth: (a) Average completion time, (b) Average response time, (c) Edge computing percentage.}
\label{f7}
\end{figure*}

\subsection{Performance with Different Numbers of UAV Devices}
In this section, we evaluate the performance of the proposed HASAC-based computation offloading and resource allocation algorithm under varying numbers of UAVs. The comparison includes five benchmark algorithms. The evaluation focuses on three metrics: average completion time, average response time, and edge computing percentage.

As shown in Fig.~\ref{f6}(a), the average completion time decreases for all algorithms as the number of UAV devices increases from 2 to 10. HASAC consistently achieves the lowest completion time across all configurations. For example, with 10 UAVs, HASAC reduces the completion time to below 1 second, whereas GCT and RO remain above 1.5 seconds. The decrease in HASAC's completion time is also steeper compared to HAA2C and CLB, reflecting its superior scalability in managing growing edge resources. PH also improves with more UAVs but lags behind the learning-based approaches.

In Fig.~\ref{f6}(b), HASAC maintains the lowest average response time across all UAV configurations. When the number of UAVs reaches 10, HASAC achieves a response time close to 0.2 seconds, which is significantly lower than that of GCT and RO, both of which exceed 0.5 second. HAA2C and CLB perform moderately well, yet their response times remain consistently higher than HASAC. RO shows the highest response delay, indicating poor adaptability to increased edge resources.

Fig.~\ref{f6}(c) shows the edge computing percentage, representing the proportion of tasks offloaded to UAVs and vessels. HASAC consistently achieves the highest offloading ratio, starting at around 55\% with 2 UAVs and increasing to over 90\% with 10 UAVs. HAA2C and CLB also improve, reaching approximately 85\% and 80\%, respectively. In contrast, PH, GCT, and RO show lower offloading ratios, with RO performing the worst across all UAV levels. This demonstrates HASAC's stronger capability in leveraging available edge computing capacity.

In summary, HASAC outperforms all baseline algorithms in completion time, response time, and edge utilization. It adapts effectively to increased UAV density, optimizes task execution, and balances computational loads efficiently. These results validate HASAC's robustness and scalability in dynamic maritime MEC environments with variable edge resource availability.

\subsection{Performance with Different MIoT Device Communication Bandwidth}

In this section, we evaluate the performance of the proposed HASAC-based computation offloading and resource allocation algorithm under varying MIoT device communication bandwidths. We compare HASAC against five benchmark algorithms. The evaluation focuses on three key metrics: average completion time, average response time, and edge computing percentage.

Fig.~\ref{f7}(a) shows that the average completion time decreases for all algorithms as the communication bandwidth increases from 1 Mbps to 20 Mbps. HASAC consistently achieves the lowest completion time across all bandwidth levels. For example, at 1 Mbps, HASAC completes tasks in approximately 5 seconds, while RO and GCT exceed 10 seconds. As bandwidth improves, HASAC's time drops below 1.5 seconds, significantly faster than all other algorithms. In contrast, RO and GCT reduce completion time more slowly due to inefficient task scheduling. HAA2C and CLB improve moderately but still fall behind HASAC, especially under high-bandwidth conditions.

Fig.~\ref{f7}(b) presents the average response time. HASAC maintains the lowest response time across all configurations, with response delay shrinking from around 1.8 seconds at 1 Mbps to less than 0.5 seconds at 20 Mbps. In comparison, RO and GCT remain above 2 seconds under low-bandwidth settings and only marginally improve at higher bandwidths. HAA2C and PH show moderate improvements, but their responsiveness does not match that of HASAC, particularly under bandwidth-constrained conditions. These results highlight HASAC's superior ability to adapt to varying communication capacities.

Fig.~\ref{f7}(c) illustrates the edge computing percentage, representing the proportion of tasks offloaded to UAVs and vessels. HASAC consistently achieves the highest offloading percentage across all bandwidth levels, increasing from approximately 50\% at 1 Mbps to nearly 90\% at 20 Mbps. HAA2C and CLB show steady growth but plateau around 75-80\%. PH, GCT, and RO lag behind significantly, with RO showing the lowest edge utilization (below 60\%) even at maximum bandwidth. This demonstrates HASAC's strong capability in leveraging high-bandwidth conditions to maximize edge resource usage.

In summary, HASAC outperforms all baseline algorithms across all bandwidth settings. It minimizes task execution time, reduces response delays, and maximizes edge computing utilization. These results confirm HASAC's robustness and adaptability in dynamic communication environments where bandwidth availability significantly affects system performance.

\section{CONCLUSIONS}\label{s7}
In this paper, we concentrate on decision-making for maritime MEC by the cooperation of UAVs and vessels. Specifically, we propose a cooperative MEC framework for MEC to optimize computation offloading and resource allocation. 
Then, the collaborative computation offloading and resource allocation problem is modeled as a MG. Furthermore, we introduce a deep reinforcement learning-based heterogeneous agent soft actor-critic algorithm to address this issue, where each agent independently generates decisions on computation offloading and resource allocation to maximize long-term rewards. 
Simulation results demonstrate that our algorithm outperforms in aspects such as convergence, execution time, computation rate, offloaded data, and the percentage of task execution across various environmental conditions, effectively generating strategies to minimize the total execution time. 
Besides, the proposed algorithm consistently achieves superior performance under diverse environmental conditions.\\

\begin{appendices}
\section{}\label{appen}
Let the real numbers $a$, $b$, and $c$ be nonnegative, $d = \max[a-b+c,0]$, then $d^2 \leq a^2+b^2+c^2+2a(c-b)$ \cite{boyd2004convex}, we have
\begin{equation}
\begin{split}
Q_i^{m}&(t+1) \leq (Q_i^{m}(t))^2+\bigg(\sum_{j = 1}^J \tau R_{i,j}^{m2u}(t)\bigg)^2\\
&+\bigg(A_i(t))^2+2Q_i^{m}(t)(A_i(t)-\sum_{j = 1}^J \tau R_{i,j}^{m2u}(t)\bigg),
\end{split}
\end{equation}
\begin{equation}
\begin{split}
&Q_j^{u}(t+1) \leq (Q_j^{u}(t))^2 +\bigg(\sum_{k = 1}^K \tau R_{j,k}^{u2v}(t)-\tau f_{i,j}^u(t)\bigg)^2\\
&+\bigg(\sum_{i = 1}^I \tau R_{i,j}^{m2u}(t)\bigg)^2+2Q_j^{u}(t)\bigg[\sum_{i = 1}^I \tau R_{i,j}^{m2u}(t)  \\
&\qquad\qquad\qquad\qquad\qquad-\bigg(\sum_{k = 1}^K \tau R_{j,k}^{u2v}(t)-\tau f_{i,j}^u(t)\bigg)\bigg],
\end{split}
\end{equation}
and
\begin{equation}
\begin{split}
Q_k^{v}(t+1) &\leq (Q_k^{v}(t))^2+(\tau f_{j,k}^v(t))^2+\bigg(\sum_{j = 1}^J \tau  R_{j,k}^{u2v}(t)\bigg)^2\\
&\qquad\qquad+2Q_k^{v}(t)\bigg(\sum_{j = 1}^J \tau R_{j,k}^{u2v}(t)-\tau f_{j,k}^v(t)\bigg).
\end{split}
\end{equation}
By substituting the above inequality and the Lyapunov function into the Lyapunov drift, we can obtain

\begin{equation}
\begin{split}
&\Delta(Q(t)) = \mathbb{E} [L(Q(t+1)) - L(Q(t))|Q(t)]\\ 
&\leq \frac{1}{2}\bigg\{\sum_{i=1}^{I}\bigg(\sum_{j = 1}^J \tau R_{i,j}^{m2u}(t)\bigg)^2+(A_i(t))^2\\
&+\sum_{j=1}^{J}\bigg(\sum_{k = 1}^K \tau  R_{j,k}^{u2v}(t)-\tau f_{i,j}^u(t)\bigg)^2+\sum_{k=1}^{K}\bigg(\tau f_{j,k}^v(t)\bigg)^2\\
&\qquad +\bigg(\sum_{i = 1}^I \tau  R_{i,j}^{m2u}(t)\bigg)^2+\bigg(\sum_{j = 1}^J\tau R_{j,k}^{u2v(t)}\bigg)^2\bigg\}\\
&+\mathbb{E} \bigg\{ \sum_{i=1}^{I} Q_i^{m}(t) \bigg( A_i(t) - \sum_{j=1}^J \tau  R_{i,j}^{m2u}(t) \bigg)  \bigg|Q(t) \bigg\}\\
&+\mathbb{E}\bigg\{\sum_{j = 1}^J Q_j^{u}(t) \bigg( \tau  R_{i,j}^{m2u}(t) \\
&\qquad\qquad\quad\ - \bigg(\sum_{k = 1}^K \tau  R_{j,k}^{u2v}(t) - \tau f_{i,j}^u(t)\bigg)\bigg) \bigg| Q(t) \bigg\}\\
&+\mathbb{E} \bigg\{ \sum_{k=1}^{K} Q_k^{v}(t)\bigg(\sum_{j = 1}^J \tau  R_{j,k}^{u2v}(t)-\tau f_{j,k}^v(t)\bigg) \bigg|Q(t) \bigg\}.
\end{split}
\end{equation}

Given the transmission rate and task arrival rate constraints, it holds that $R_{i,j}^{m2u}(t) \leq R_j^{umax}$, $R_{j,k}^{u2v}(t)\leq R_k^{vmax}$, and $A_i(t) \leq A_i^{max}$. Therefore, we can define the following equation
\begin{equation}
\begin{split}
D = \frac{1}{2} & \bigg\{\sum_{i=1}^{I}[\tau R_j^{umax} + (A_i^{max})^2] + \sum_{j=1}^{J}[(\tau R_k^{vmax} - \tau f_j^{max})^2\\
&\qquad + (\tau R_j^{umax})^2]+ \sum_{k=1}^{K} [ (\tau R_v^{umax})^2 - (\tau f_k^{vmax} )^2 ] \bigg\}.
\end{split}
\end{equation}

\end{appendices}

\bibliography{references}
\end{document}